\documentclass[letterpaper, 10 pt, conference]{ieeeconf}  

\IEEEoverridecommandlockouts                              

\usepackage{graphics} 
\usepackage{float} 
\usepackage{epsfig} 
\usepackage{mathptmx} 
\usepackage{times} 
\usepackage{amsmath} 
\usepackage{amssymb}  
\usepackage{tikz}
\usepackage{tikz-3dplot}
\usepackage{relsize}
\usepackage{multirow}
\usepackage{graphicx}
\usepackage{subfigure}
\usepackage{cite}
\usepackage{balance}
\usepackage[table]{xcolor}

\title{\LARGE \bf
A Height-Constrained 2-Point Minimal Solver for Pose Estimation\\ from Active LED Markers with Event Cameras
}

\author{Runze Yuan$^{1}$, Alexander Kappler$^{1}$, Jun Zhang$^{1}$, Kuangyi Chen$^{1}$, Fabio Morbidi$^{2}$, \\ Pascal Vasseur$^{2}$, Cédric Demonceaux$^{3}$, and Friedrich Fraundorfer$^{1}$ 
\thanks{$^{1}$ Institute of Visual Computing, Graz University of Technology}%
\thanks{$^{2}$ MIS laboratory, University of Picardie Jules Verne}%
\thanks{$^{3}$ ICB laboratory, University of Burgundy Europe}%
}

\begin{document}
\setlength{\textfloatsep}{4pt}
\setlength{\dbltextfloatsep}{4pt}

\maketitle
\thispagestyle{empty}
\pagestyle{empty}

\begin{abstract}
In many autonomous applications requiring real-time localization, active marker-based systems are preferred due to their low latency and ease of deployment compared to computationally demanding feature-based methods. Event~\mbox{cameras} offer high temporal resolution and minimal delay and are commonly used with active LED markers for robust real-time localization.
Existing methods typically rely on Perspective-n-Point (PnP) solvers for pose estimation. However, structured marker layouts can be challenging to deploy in space-constrained scenarios, while partial self-motion information (e.g., gravity direction and altitude) is readily available from onboard sensors. We derive a robust and accurate minimal solver that estimates camera pose from only two LED markers by incorporating known tilt angle and camera height measured by an onboard sensor, such as an IMU or an altimeter. The proposed formulation uniquely determines the camera pose through both a closed-form and a linear least-squares solution. We further analyze degenerate configurations and characterize the conditions under which height information does not contribute to rotation estimation. For evaluation, we developed an event-based active marker system to collect real-world data with ground truth from a motion capture system. Experiments on both synthetic and real data demonstrate improved accuracy over the state-of-the-art P2P solver and competitive performance relative to P3P.

\end{abstract}


\section{INTRODUCTION}

Fast and accurate camera pose estimation is essential for vision-based robotic systems operating in GNSS-denied environments, such as indoor spaces or underground tunnels. In most of the cases, 
PnP solvers~\cite{lepetit2009ep,hesch2011direct,kneip2014upnp} are utilized as standard techniques to determine the 6-DoF (Degrees of Freedom) camera pose relative to a known 3D map, which requires both prior map creation and maintenance. They are frequently used in settings where a large number of 3D-2D correspondences are available by feature matching, between the map and the current view. In these conditions, feature detection, matching, and outlier rejection are necessary but computationally expensive operations, which can degrade real-time performance.

In contrast, infrastructure-based localization like marker-based methods (e.g., AprilTag, ArUco, or active LED \mbox{markers}), avoids map building in advance and achieves pose estimation using only a few 3D-2D matches.
Under these provisions, pose estimation algorithms are supposed to work with the minimal number of matches required, to keep the number of markers low, e.g., to facilitate \mbox{practical} \mbox{deployment}. It~is~therefore important to devise pose \mbox{estimation} algorithms that work well in practice and that are accurate with the minimal number of data points.

Traditional marker-based methods rely on standard optical cameras to track markers at constant frame rates. Such~an operating mode limits their performance when handling scenarios with fast motion and variable light conditions. \emph{Event~cameras} are neuromorphic sensors inspired by biological vision, which record light-intensity changes asynchronously, at individual pixels. This mechanism guarantees high temporal resolution and a significantly expanded dynamic range, compared to conventional cameras. These properties make them ideal for low-latency tracking of actively blinking markers for camera localization, even under challenging ambient light conditions~\cite{tofighi2025survey}. 

\begin{figure}[t!]
    \centering
    \includegraphics[width=\linewidth]{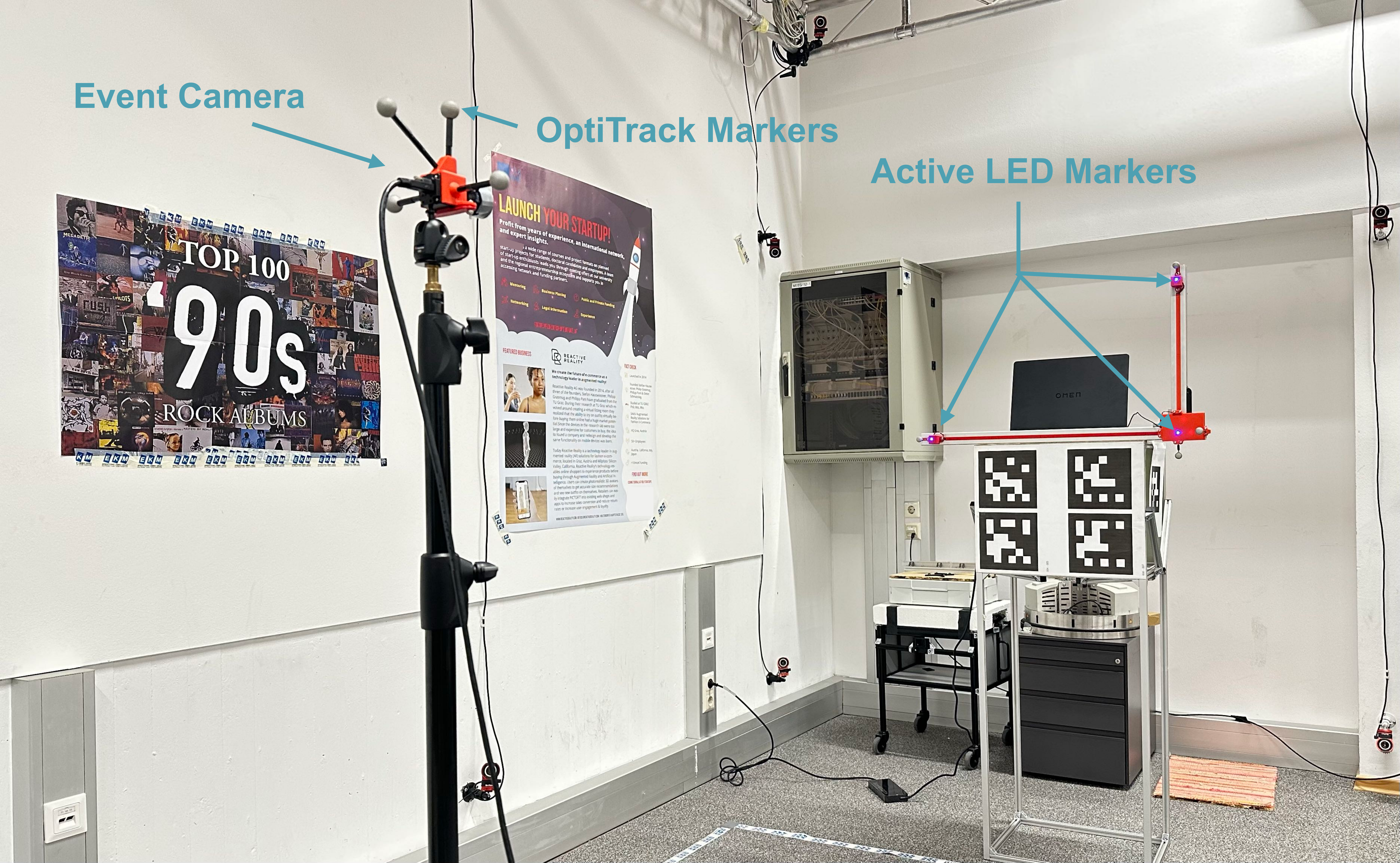}
    \caption{\emph{Overview of our experimental setup}. Three active LED markers are fixed on a rig with known 3D positions from an OptiTrack system. A~freely moving event camera is used to estimate its 3-DoF pose from the detected LED markers, given the height and tilt angle as priors. Note that our solution only requires two markers: the third one is utilized for the P3P solver, as a baseline.}
    \label{fig:teaser}
\end{figure}

Motivated by the previous observations, this paper \mbox{develops} a robust and accurate minimal solver for marker-based pose estimation with an event camera. To this end, we design an event-based active marker system with two LED light sources that are programmable in frequency for detection, 
and a simple but effective marker detection pipeline from the event stream (see Fig.~\ref{fig:teaser}). On this basis, we formulate the P2P problem with 

known vertical direction and camera height. In the majority of mobile robots, such as ground or aerial vehicles, these two quantities can be reliably obtained from onboard sensors (e.g., IMUs or altimeters).
We~provide two solutions to determine the camera pose: a closed-form and a linearized least-squares solution. Compared to the state-of-the-art P2P~\cite{li2023generalized} and P3P~\cite{kneip2011novel} solvers, our P2P solver returns a unique solution without requiring extra points to resolve pose ambiguity, and achieves better accuracy in deriving the position of the camera, especially under larger viewing distances from the markers.

In conclusion, the main contributions of this paper can be summarized as follows:
\begin{itemize}

    \item We derive a robust, accurate, and ambiguity-free P2P solver incorporating tilt angle and height priors, admitting both a closed-form solution and a linearized least-squares formulation. In addition, we present a comprehensive theoretical analysis of the proposed solver, identifying degenerate configurations and conditions where the height prior does not influence rotation estimation.
    \item We propose an event-based active marker system to \mbox{facilitate} the development and benchmarking of minimal-point solvers for precise pose estimation with event cameras.
    \item We conduct extensive evaluations on synthetic and real data, demonstrating the effectiveness and advantage of our proposed minimal solver against the state-of-the-art P2P and P3P solvers.
\end{itemize}

The remainder of this article is organized as follows. In~Sect.~\ref{Sect:Rel-Work}, the related work is presented, while in Sect.~\ref{Sect:Prob-Form} the problem studied in the paper is formulated. The results of synthetic and real-world experiments are discussed in Sect.~\ref{Sect:Exper}. Finally, conclusions are drawn in Sect.~\ref{Sect:Concl}.

\section{RELATED WORK}\label{Sect:Rel-Work}

\subsubsection{Event-based Active Marker Systems}

Marker-based localization systems using event cameras have made steady progress in recent years~\cite{censi2013low,jianhong2019method,chen2020novel,salah2022neuromorphic,ebmer2024real,bauersfeld2025monocular}. Censi \emph{et al.} \cite{censi2013low} presented the first pose tracking system based on an event camera with active LED markers and formulate LED tracking as a probabilistic inference problem, demonstrating robust performance under fast and aggressive motions. In contrast, Xu \emph{et al.} \cite{jianhong2019method} employed a frequency-filtering and clustering pipeline followed by a PnP solver, and validated their method in low-light conditions. Inspired by \cite{censi2013low}, Chen \emph{et al.} \cite{chen2020novel} further incorporated a Gaussian mixture probability hypothesis density (GM-PHD) filter for multi-LED tracking, achieving high accuracy indoor localization while avoiding conventional image processing. Salah \emph{et al.}~\cite{salah2022neuromorphic} subsequently developed an event-based relative localization framework between a drone and a ground vehicle, fusing active LED observations with inertial measurements for enhanced robustness.
Later, Ebmer \emph{et al.} \cite{ebmer2024real} proposed a low-latency localization system that exploits bias and timing priors to further reduce estimation delay. A more recent development \cite{bauersfeld2025monocular} further refines detection and fusion strategies toward a monocular event-based motion capture system. It is worth noting that most marker-based methods employ PnP solvers with at least three point correspondences for pose estimation. Such configurations typically require multiple active LED markers, which increase power consumption and hardware complexity in practical deployments.

\subsubsection{Existing 2-point and 3-point Algorithms}

Modern P3P algorithms~\cite{gao2003complete, kneip2011novel,ke2017efficient,persson2018lambda} primarily focus on improving numerical stability and computational efficiency by utilizing closed-form algebraic solutions, instead of the classical geometric method~\cite{grunert1841pothenotische}. As an example, a direct parametrization is proposed in~\cite{kneip2011novel} to compute the camera position and orientation in a unique coordinate frame, avoiding intermediate distance calculations, hence reducing computational overhead. 
Instead of solving a quartic equation, in~\cite{persson2018lambda} the authors take advantage of a unique ``twist'' transformation and a small system of equations to enhance accuracy in degenerate configurations. Note that all these P3P algorithms generally return up to four possible solutions, and a fourth point is usually required to determine the unique solution.

A P2P solver can further reduce computational complexity by using only two points, which is possible when additional information is available, such as known vertical direction or motion prior~\cite{troiani20142, choi2015new, sweeneySolver, gao20172, li2023generalized}. For instance, an efficient two-point solver is proposed in~\cite{sweeneySolver} to compute the absolute pose of a single or multi-camera system with knowledge of vertical direction. In~\cite{li2023generalized}, the authors further improved the numerical stability and accuracy by introducing a quadratic polynomial involving only one variable about the orientation from the geometric constraints. However, such two-point solvers generally yield two feasible solutions, requiring an additional point correspondence to resolve the ambiguity, which consequently increases the hardware cost and system complexity of marker-based setups.

Motivated by the availability of height measurements in robotic systems, we reformulate the P2P problem with an additional height prior to better constrain the camera position and improve robustness, particularly at large viewing distances. In contrast to existing marker-based localization systems, the proposed solver requires only two point correspondences and does not rely on an additional point to resolve solution ambiguity.

\section{PROBLEM FORMULATION}\label{Sect:Prob-Form} 

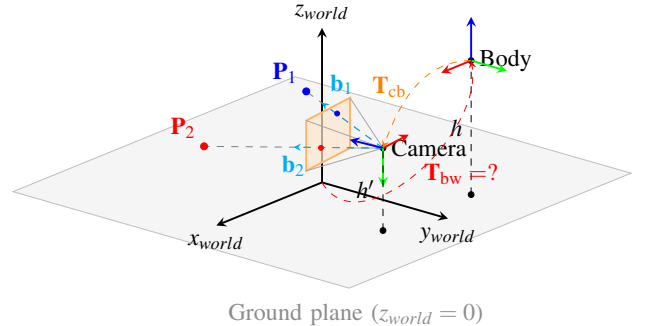
\begin{figure}[b!]
  \centering
  \resizebox{\columnwidth}{!}{\tdplotsetmaincoords{70}{130} 
\begin{tikzpicture}[tdplot_main_coords,scale=0.8,>=stealth]
  \fill[gray!8] (-4,-4,0) -- (4,-4,0) -- (4,4,0) -- (-4,4,0) -- cycle;
  \draw[gray!50] (-4,-4,0) -- (4,-4,0) -- (4,4,0) -- (-4,4,0) -- cycle;
  \node[gray!90] at (5,5,0) {Ground plane ($z_{world}=0$)};
  \coordinate (W0) at (0,0,0);
  \draw[->,thick] (W0) -- (3,0,0) node[below] {$x_{world}$};
  \draw[->,thick] (W0) -- (0,3,0) node[below] {$y_{world}$};
  \draw[->,thick] (W0) -- (0,0,3) node[above] {$z_{world}$};

  \pgfmathsetmacro{\hh}{1.6}

  \coordinate (P0) at (0.15,-0.25,\hh+0.15);  
  \coordinate (Ph) at (2.75,-0.5,\hh-0.25);  

  \coordinate (D) at (1.25,2.5,\hh); 
  \filldraw[black] (D) circle (1.5pt) node[right] {Camera};

  \coordinate (zCamDir) at (0.05,-0.75,0);        
  \coordinate (yCamDir) at (0,0,-0.75);       
  \coordinate (xCamDir) at (-0.85,-0.1,0);        

  \pgfmathsetmacro{\ipl}{1.65}  
  \pgfmathsetmacro{\iw}{0.85}   
  \pgfmathsetmacro{\ih}{0.65}   

  \coordinate (Icenter) at ($(D)+\ipl*(zCamDir)$);

  \coordinate (I1) at ($(Icenter) + \iw*(xCamDir) - \ih*(yCamDir)$);
  \coordinate (I2) at ($(Icenter) - \iw*(xCamDir) - \ih*(yCamDir)$);
  \coordinate (I3) at ($(Icenter) - \iw*(xCamDir) + \ih*(yCamDir)$);
  \coordinate (I4) at ($(Icenter) + \iw*(xCamDir) + \ih*(yCamDir)$);

  \fill[orange!20,opacity=0.7] (I1) -- (I2) -- (I3) -- (I4) -- cycle;
  \draw[orange!60,thick] (I1) -- (I2) -- (I3) -- (I4) -- cycle;

  \draw[gray!70] (D) -- (I1);
  \draw[gray!70] (D) -- (I2);
  \draw[gray!70] (D) -- (I3);
  \draw[gray!70] (D) -- (I4);

  \draw[dashed,gray] (D) -- (P0);
  \draw[dashed,gray] (D) -- (Ph);

  \draw[->, dashed, cyan] (D) -- ($(D)!0.8!(P0)$) node[above right] {$\mathbf{b}_{1}$};;
  \draw[->, dashed, cyan] (D) -- ($(D)!0.5!(Ph)$) node[below] {$\mathbf{b}_{2}$};;

  \filldraw[blue] (P0) circle (1.8pt) node[above left] {$\mathbf{P}_{\text{1}}$};
  \filldraw[red]  (Ph) circle (1.8pt) node[above left] {$\mathbf{P}_{\text{2}}$};

  \filldraw[red]  (1.4,1.15,1.33) circle (1.25pt);
  \filldraw[blue]  (1.06,1.25,1.925) circle (1.25pt);

  \draw[->,thick,blue]  (D) -- ($(D)+(zCamDir)$) node[above right] {};
  \draw[->,thick,green] (D) -- ($(D)+(yCamDir)$) node[right] {};
  \draw[->,thick,red]   (D) -- ($(D)+(xCamDir)$) node[right] {};

    \coordinate (Ddrone) at ($(D)+(-2.5,-0, 1)$);
    \filldraw[black] (Ddrone) circle (1.5pt) node[right] {Body};
    
    \coordinate (xDroneDir) at (0.85,0,0);
    \coordinate (yDroneDir) at (0,0.85,0);
    \coordinate (zDroneDir) at (0,0,0.85);
    
    \draw[->,thick,red]   (Ddrone) -- ($(Ddrone)+(xDroneDir)$) node[below right] {};
    \draw[->,thick,green] (Ddrone) -- ($(Ddrone)+(yDroneDir)$) node[left]        {};
    \draw[->,thick,blue]  (Ddrone) -- ($(Ddrone)+(zDroneDir)$) node[above]       {};
    
    
    \draw[->,dashed,orange]
      (Ddrone) to[out=180,in=70]
      node[midway, left, text=orange] {$\mathbf{T}_{\mathrm{cb}}$}
      (D);
      
  \draw[dashed] (-1.25,2.5,0) -- (Ddrone) node[midway,left] {$h$};
    \filldraw[black] (-1.25,2.5,0) circle (1.5pt);
    
    \draw[->,dashed,red]
      ($(0,0,0)$) to[out=-60,in=-80]
      node[midway, below, right, text=red] {$\mathbf{T}_{\mathrm{bw}} = ?$}
      (Ddrone);

      \draw[dashed] (1.25,2.5,0) -- (D) node[midway,left] {$h'$};
        \filldraw[black] (1.25,2.5,0) circle (1.5pt);
\end{tikzpicture}}
  \caption{\emph{Graphical illustration of the problem studied in Sect.~\ref{Sect:Prob-Form}}. The red, green, and blue axes denote the $x$-, $y$-, and $z$-axes, respectively. 
  } 
  \label{fig:problem}
\end{figure}

In this section, we consider the problem shown in Fig.~\ref{fig:problem}. An event camera is attached to the body frame of a robot (e.g., a drone) via a known extrinsic transformation $\mathbf{T}_{\mathrm{cb}} = \left[\begin{smallmatrix}
    \mathbf{R}_{\mathrm{cb}} & \mathbf{t}_{\mathrm{cb}}\vspace{0.05cm}\\
    \mathbf{0} & 1 
\end{smallmatrix}\right]$.
The vertical direction $\mathbf{d}_{\mathrm{up}}$ and altitude $h$ are observable from onboard sensors (e.g., IMU and altimeter). Two LED markers, $\mathbf{P}_1$ and $\mathbf{P}_2$, are located in the world frame and are observable by the event camera. Their spatial coordinates $\mathbf{P}_1 = [\mathrm{X}_1,\, \mathrm{Y}_1,\, Z_1]^\top$, $\mathbf{P}_2 = [\mathrm{X}_2,\, \mathrm{Y}_2,\, \mathrm{Z}_2]^\top$ are assumed to be known.
The objective is to find the Euclidean transformation from the world frame to the body frame, $\mathbf{T}_{\mathrm{bw}} = \left[\begin{smallmatrix}
    \mathbf{R}_{\mathrm{bw}} & \mathbf{t}_{\mathrm{bw}}\vspace{0.05cm}\\
    \mathbf{0} & 1 
\end{smallmatrix}\right]$.

We parameterize the rotation using the Euler angles, $\mathbf{R}_{\mathrm{bw}} = \mathbf{R}_x\mathbf{R}_y\mathbf{R}_z$, where $\mathbf{R}_x$, $\mathbf{R}_y$ and $\mathbf{R}_z$ are the elementary rotation matrices about the $x$-, $y$-, and $z$-axes, respectively. Considering the calibrated perspective camera model with intrinsic matrix $\mathbf{K} \in \mathbb{R}^{3 \times 3}$, each image measurement $\mathbf{x}_i$ could be expressed as a unit vector pointing from the camera center to the marker,
i.e., the bearing vector $\mathbf{b}_i \in \mathbb{R}^3$~\cite{opengv} given by, 
$$
    \mathbf{b}_i \,=\, \frac{\mathbf{K}^{-1}\mathbf{x}_i}{\|\mathbf{K}^{-1}\mathbf{x}_i\|} \,\triangleq\, s_i\!\begin{bmatrix}
        u_i \\
        v_i \\
        1
    \end{bmatrix}\!, 
$$
where $s_i$ is a positive scale factor and ``$\triangleq$'' denotes equality by definition. 
%
In the general case, we can enforce the point-ray collinearity constraint,
\begin{equation}
    \mathbf{b}_i \times (\mathbf{R}_{\mathrm{cb}}\mathbf{R}_x\mathbf{R}_y \mathbf{R}_z \mathbf{P}_i + \mathbf{R}_{\mathrm{cb}} \mathbf{t}_{\mathrm{bw}} + \mathbf{t}_{\mathrm{cb}}) \,=\, \mathbf{0}, \quad i \in\{1, 2\},
    \label{eq:general_case}
\end{equation}
where ``$\times$'' denotes the cross product.
With the knowledge of the vertical direction (e.g., measured by an IMU), the tilt component $\mathbf{R}_x\mathbf{R}_y$ can be solved in closed form via 
\begin{equation*}
    \mathbf{R}_x\mathbf{R}_y\mathbf{d}_{\mathrm{up}} \,=\, \mathbf{e}_z,
    \label{eq:vertical}
\end{equation*}
where $\mathbf{e}_z = [0,\, 0,\, 1]^\top$.  Given the tilt rotation and the extrinsic calibration, Eq.~\eqref{eq:general_case} can be recast as follows:
\begin{equation}
    (\mathbf{R}_y^\top\mathbf{R}_x^\top\mathbf{R}_{\mathrm{cb}}^\top\mathbf{b}_i)\times(\mathbf{R}_z\mathbf{P}_i + \mathbf{R}_y^\top\mathbf{R}_x^\top\mathbf{t}_{\mathrm{bw}} + \mathbf{R}_y^\top\mathbf{R}_x^\top\mathbf{R}_{\mathrm{cb}}^\top\mathbf{t}_{\mathrm{cb}}) \,=\, \mathbf{0}.
    \label{eq:tilt_compensated}
\end{equation}
In Eq.~\eqref{eq:tilt_compensated}, we note that:

\begin{enumerate}
    \item $\mathbf{b}_i' \,\triangleq\, \mathbf{R}_y^\top\mathbf{R}_x^\top\mathbf{R}_{\mathrm{cb}}^\top\mathbf{b}_i $ is the bearing vector expressed in the tilt-compensated robot body frame.
    
    \item $\mathbf{t}' \,\triangleq\, \mathbf{R}_y^\top\mathbf{R}_x^\top\mathbf{t}_{\mathrm{bw}} + \mathbf{R}_y^\top\mathbf{R}_x^\top\mathbf{R}_{\mathrm{cb}}^\top\mathbf{t}_{\mathrm{cb}}$ is the sum of the translation vector of the tilt-compensated body frame and a constant tilt-compensated extrinsic offset, which will be referred to as $\mathbf{t}_\Delta$. 
\end{enumerate}
We explicitly define the remaining unknowns $\mathbf{R}_z$ and $\mathbf{t'}$~as
\begin{equation}
    \mathbf{R}_z = \begin{bmatrix}
        \cos\theta & -\sin\theta & 0 \\
        \sin\theta & \cos\theta & 0 \\
        0 & 0 & 1
    \end{bmatrix}\!,\; 
    \mathbf{t}' = \begin{bmatrix}
        t_x + t_{\Delta_x}\\
        t_y + t_{\Delta_y} \\
        -h + t_{\Delta_z}
    \end{bmatrix} \triangleq 
    \begin{bmatrix}
        t_x' \\ t_y' \\ -h'
    \end{bmatrix}\!.
    \label{eq:parameter}
\end{equation}
In particular, the $z$-component of vector $\mathbf{t}'$ is simply the negative robot's height plus an offset, which remains constant over time. Consequently, the parameter vector to be solved for is just $[\theta,\, t_x',\, t_y']^\top$. Eq.~\eqref{eq:tilt_compensated} can be further simplified, leading to:
\begin{equation}
    [\mathbf{b}_i']_\times(\mathbf{R}_z\mathbf{P}_i \,+\,\mathbf{t}') \,=\, \mathbf{0},
    \label{eq:simplified}
\end{equation}
where $[\mathbf{b}_i']_\times$ is the skew-symmetric matrix associated with~$\mathbf{b}_i'$.
Substituting Eq.~\eqref{eq:parameter} into Eq.~\eqref{eq:simplified} and expanding yields:
\begin{equation}
\!\!\scalebox{0.87}{
$
    \begin{bmatrix}
        0 & -s_i' & s_i'v_i' \\
        s_i' & 0 & -s_i'u_i' \\
        -s_iv_i' & s_iu_i' & 0
    \end{bmatrix}
    \!\left(\begin{bmatrix}
        \cos\theta & -\sin\theta & 0 \\
        \sin\theta & \cos\theta & 0 \\
        0 & 0 & 1
    \end{bmatrix}
    \!
    \begin{bmatrix}
        \mathrm{X}_i \\ \mathrm{Y}_i \\ \mathrm{Z}_i
    \end{bmatrix}
    +\begin{bmatrix}
        t_x' \\ t_y' \\ -h'
    \end{bmatrix}\right) \,=\, \mathbf{0}.
$
}
\label{eq:matrix_constraint}
\end{equation}
Although Eq.~\eqref{eq:matrix_constraint} consists of three rows, it only yields two independent equations, since the skew-symmetric matrix $[\mathbf{b}_i']_\times$ is rank deficient. 
By discarding the redundant equation, 
We obtain four independent equations from two point correspondences, i.e., the system 

\begin{equation}
    \mathbf{A}\mathbf{x} \,=\, \mathbf{b},
    \label{eq:linear_constraint}
\end{equation}
where
\begin{equation*}
    \mathbf{A} =     \begin{bmatrix}
        \mathrm{X}_1& -\mathrm{Y}_1& 1& 0 \\
        \mathrm{Y}_1& \mathrm{X}_1 & 0 &1 \\
        \mathrm{X}_2& -\mathrm{Y}_2& 1& 0 \\
        \mathrm{Y}_2& \mathrm{X}_2 & 0 &1
    \end{bmatrix}\!,\, \mathbf{x} = \begin{bmatrix}
     \cos\theta \\
     \sin\theta \\
     t_x' \\
     t_y'
    \end{bmatrix}\!,\, \mathbf{b} =     \begin{bmatrix}
        u_1'(\mathrm{Z}_1 - h') \\
        v_1'(\mathrm{Z}_1 - h') \\
        u_2'(\mathrm{Z}_2 - h') \\
        v_2'(\mathrm{Z}_2 - h') \\
    \end{bmatrix}\!.
\end{equation*}

By incorporating the trigonometric-identity constraint (i.e. $\cos^2\theta + \sin^2\theta = 1$), system~\eqref{eq:linear_constraint} becomes overdetermined (in~fact, it comprises five constraints over three unknowns). Different from~\cite{li2023generalized,sweeneySolver}, which rely on an additional point correspondence to resolve pose ambiguity, the proposed formulation incorporates a height prior to yield a unique solution. This property is particularly useful for active marker–based pose estimation. In the next section, we will present two alternative solution strategies.


\subsection{Closed-Form Solution}

We first derive a closed-form solution to system~\eqref{eq:linear_constraint}.
Under the trigonometric-identity constraint, the rotation angle can be estimated independently of the translation. Specifically, we can eliminate $t_x'$ by combining the first and third row of Eq.~\eqref{eq:linear_constraint}, and we can eliminate $t_y'$ by combining the second and fourth row, yielding
\begin{align}
    \Delta \mathrm{X}\cos\theta - \Delta \mathrm{Y}\sin\theta \,=\, u_1'(\mathrm{Z}_1 - h') - u_2'(\mathrm{Z}_2 - h'), \label{eq:poly1} \\
    \Delta \mathrm{Y}\cos\theta + \Delta \mathrm{X}\sin\theta \,=\, v_1'(\mathrm{Z}_1 - h') - v_2'(\mathrm{Z}_2 - h'), \label{eq:poly2}
\end{align}
where $\Delta \mathrm{X} = \mathrm{X}_1 - \mathrm{X}_2$ and $\Delta \mathrm{Y} = \mathrm{Y}_1 - \mathrm{Y}_2$. Following the substitution proposed in~\cite{li2023generalized}, we can drop the trigonometric-identity constraint by introducing the reparameterization $s = \tan(\theta/2)$, under which the sine and cosine terms can be rewritten as
\begin{equation}
    \sin\theta \,=\, \frac{2s}{1 + s^2},  \quad \cos\theta \,=\, \frac{1 - s^2}{1+s^2}.
    \label{eq:tan}
\end{equation}
Plugging~\eqref{eq:tan} into \eqref{eq:poly1} and \eqref{eq:poly2}, yields two independent quadratic equations in $s$,
\begin{equation}
     a_is^2 \,+\, b_is \,+\, c_i \,=\, 0, \quad i \in \{1, 2\},
     \label{eq:quad}
\end{equation}
whose coefficients are given by
\begin{align*}
    a_1 &\,=\, \alpha_1 + \Delta \mathrm{X} , &b_1 &\,=\, 2\Delta \mathrm{Y} , &c_1 &\,=\,  \alpha_1 - \Delta \mathrm{X},  \\
    a_2 &\,=\, \alpha_2 + \Delta \mathrm{Y}, &b_2 &\,=\, -2\Delta \mathrm{X}, &c_2 &\,=\, \alpha_2 - \Delta \mathrm{Y},
\end{align*}
with
\begin{align*}
    \alpha_1 &\,\triangleq\, u'_1(\mathrm{Z}_1 - h') - u'_2(\mathrm{Z}_2 - h'),\\
    \alpha_2 &\,\triangleq\, v'_1(\mathrm{Z}_1 - h') - v'_2(\mathrm{Z}_2 - h').
\end{align*}
The closed-form solutions of Eqs.~\eqref{eq:quad} are given by
\begin{equation*}
    s_i \,=\, \frac{-b_i \,\pm\, 2\sqrt{\Delta \mathrm{X}^2 + \Delta \mathrm{Y}^2 - \alpha_i^2}}{2a_i}, \quad i \in \{1, 2\}. 
    \label{eq:solution}
\end{equation*}
A non-trivial implication of Eqs.~\eqref{eq:quad} is that, given a height prior, the rotation angle can be recovered \emph{solely} from the $x$- or $y$-components of the 2D measurements. 

In practice, however, it is not immediately clear which of the two quadratic equations in~\eqref{eq:quad} should be used to compute the solution. From a numerical standpoint, the system becomes severely ill-conditioned as the discriminant approaches zero \cite{higham2002accuracy}, i.e., when $\Delta \mathrm{X}^2 + \Delta \mathrm{Y}^2 - \alpha_i^2 \approx 0$.  Since $\Delta \mathrm{X}^2 +\Delta \mathrm{Y}^2$ is constant in our setting, we adopt a simple yet effective discriminant-based selection criterion: we choose the quadratic equation associated with the smaller value of $|\alpha_i|$, which corresponds to a better-conditioned formulation. The~remaining equation then serves as an independent constraint for resolving the ambiguity, yielding a unique solution. Once the rotation angle is determined, the translation vector can be recovered by substituting it into Eq.~\eqref{eq:linear_constraint}.


\subsection{Linear Least-Squares Solution} 

With four equations available for 3~DoF, a least-squares (LS) solution to system~\eqref{eq:linear_constraint} can be found,
while temporarily ignoring the trigonometric-identity constraint. The~system can then be solved via Singular Value Decomposition (SVD),
\begin{equation*}
    \mathbf{x} \,=\, \mathbf{A}^{+}\,\mathbf{b},
\end{equation*}
where $\mathbf{A}^{+}$ denotes the Moore–Penrose pseudoinverse. After back-substituting $\mathbf{x}$ using Eq.~\eqref{eq:parameter}, the resulting $\mathbf{\tilde{R}}_z$ is not necessarily a valid rotation matrix and it must be mapped back onto $\mathrm{SO}(3)$ by performing an SVD on $\mathbf{\tilde{R}}_z = \mathbf{U}\Sigma\mathbf{V}^\top$ and setting $\mathbf{R}_z = \mathbf{U}\mathbf{V}^\top$. 

\subsection{Theoretical Analysis of the Proposed Solver}\label{Sect:TheorAnal}

Like other minimal solvers, our approach also suffers from certain limitations. Therefore, we present here a comprehensive analysis of the proposed solver under degenerate and special cases. In particular, we analyze degenerate configurations in which the formulation either fails to yield a meaningful solution or becomes highly sensitive to noise. In addition, we consider a special case where the height measurement does not contribute to rotation estimation, under which the proposed solver reduces to the method presented in~\cite{li2023generalized}. Geometrically, degeneracy can arise either from the spatial configuration of the observed 3D points or from the camera motion itself. We thereby analyze cases from these two complementary perspectives.

\subsubsection{Spatial Configuration of Two Points}
\begin{itemize}
    \item A simple degenerate case arises from the structure of matrix $\mathbf{A}$ in Eq.~\eqref{eq:linear_constraint}. Specifically, the determinant of $\mathbf{A}$ is given by
    \[
        \det(\mathbf{A}) \,=\, (\mathrm{X}_1 - \mathrm{X}_2)^2 \,+\, (\mathrm{Y}_1 - \mathrm{Y}_2)^2.
    \]
    It follows that $\det(\mathbf{A}) = 0$ if and only if $(\mathrm{X}_1, \mathrm{Y}_1) = (\mathrm{X}_2, \mathrm{Y}_2)$. 
    Geometrically, this corresponds to the case where the two points $\mathbf{P}_1$ and $\mathbf{P}_2$ have identical planar coordinates and differ only along the vertical direction. 

    \item Another special configuration arises when the points $\mathbf{P}_1$ and $\mathbf{P}_2$ have identical heights, i.e. $\mathrm{Z} \triangleq \mathrm{Z}_1 = \mathrm{Z}_2$. In~this case, system~\eqref{eq:linear_constraint} can be written as 
    \[
        \mathbf{A}\mathrm{x} \,=\, \lambda \mathbf{b}',
    \]
    where $\lambda = \mathrm{Z} - h'$ is a scalar factor and $\mathbf{b}' = [u'_1,\, v'_1,\, u'_2,\, v'_2]^\top$. Since $\mathbf{b}'$ is independent of the height prior, the height information collapses into a scale factor~$\lambda$, which is subsequently removed during the projection onto $\mathrm{SO}(3)$. Consequently, the rotation estimation relies solely on two-point correspondences without height constraints and reduces to the formulation in~\cite{li2023generalized}. Importantly, this behavior is specific to the linear LS formulation. The closed-form derivation analytically eliminates translation and preserves the influence of the height prior on rotation estimation. Therefore, the height information continues to contribute to rotation estimation even when $\mathbf{P}_1$ and $\mathbf{P}_2$ share the same height.
\end{itemize}

\subsubsection{Camera Motion}

\begin{itemize}
    \item  Certain camera motions can also lead to degenerate solutions or ill-conditioned estimation problems. In particular, when the camera is located at the same height as one of the two points, i.e., when $\mathrm{Z}_i - h' = 0$ for $i \in \{1,2\}$, the corresponding two rows of the observation vector $\mathbf{b}$ in Eq.~\eqref{eq:linear_constraint} vanish. As a result, the constraints associated with that point no longer provide useful information.
    In~the extreme case where the height of $\mathbf{P}_1$, $\mathbf{P}_2$ and the camera is the same height, $\mathbf{b} = \mathbf{0}$, and the linear system only admits the trivial solution $\mathbf{x}= \mathbf{0}$.
\end{itemize}

The previous analysis indicates that both point selection and camera motion play a critical role in ensuring observability. From a practical point of view, degenerate cases can be avoided by selecting 3D points with sufficient horizontal separation and height diversity. Moreover, the camera should maintain an adequate height difference with respect to the observed points. For example, in aerial robotics, placing the two points on non-planar terrain naturally satisfies this requirement.


\section{EXPERIMENTS}\label{Sect:Exper}

This section presents the experimental evaluation of the proposed P2P solver.
We first conduct a synthetic analysis to evaluate the robustness of the solver under varying pixel noise levels and target distances. The degenerate configurations identified in Sect.~\ref{Sect:TheorAnal} are also explicitly evaluated to validate our findings. 
We then present the real-world experimental setup, including the hardware configuration and the 2D keypoint detection pipeline. Finally, real-world experiments are conducted to demonstrate the practical applicability of the proposed approach. For comparison, we include the P3P solver of Kneip \emph{et al.}~\cite{kneip2011novel}, and the P2P solver proposed by Li \emph{et al.}~\cite{li2023generalized} as baselines. Both the closed-form and the LS of our method are considered in our comparative analysis.
 
\subsection{Analysis on Synthetic Data}

\begin{figure*}[t!]
    \centering
    \vspace{5pt}
    \subfigure[Median translation (m) and rotation (deg) errors under varying pixel noise levels.]{ 
            \includegraphics[width=0.49\linewidth]{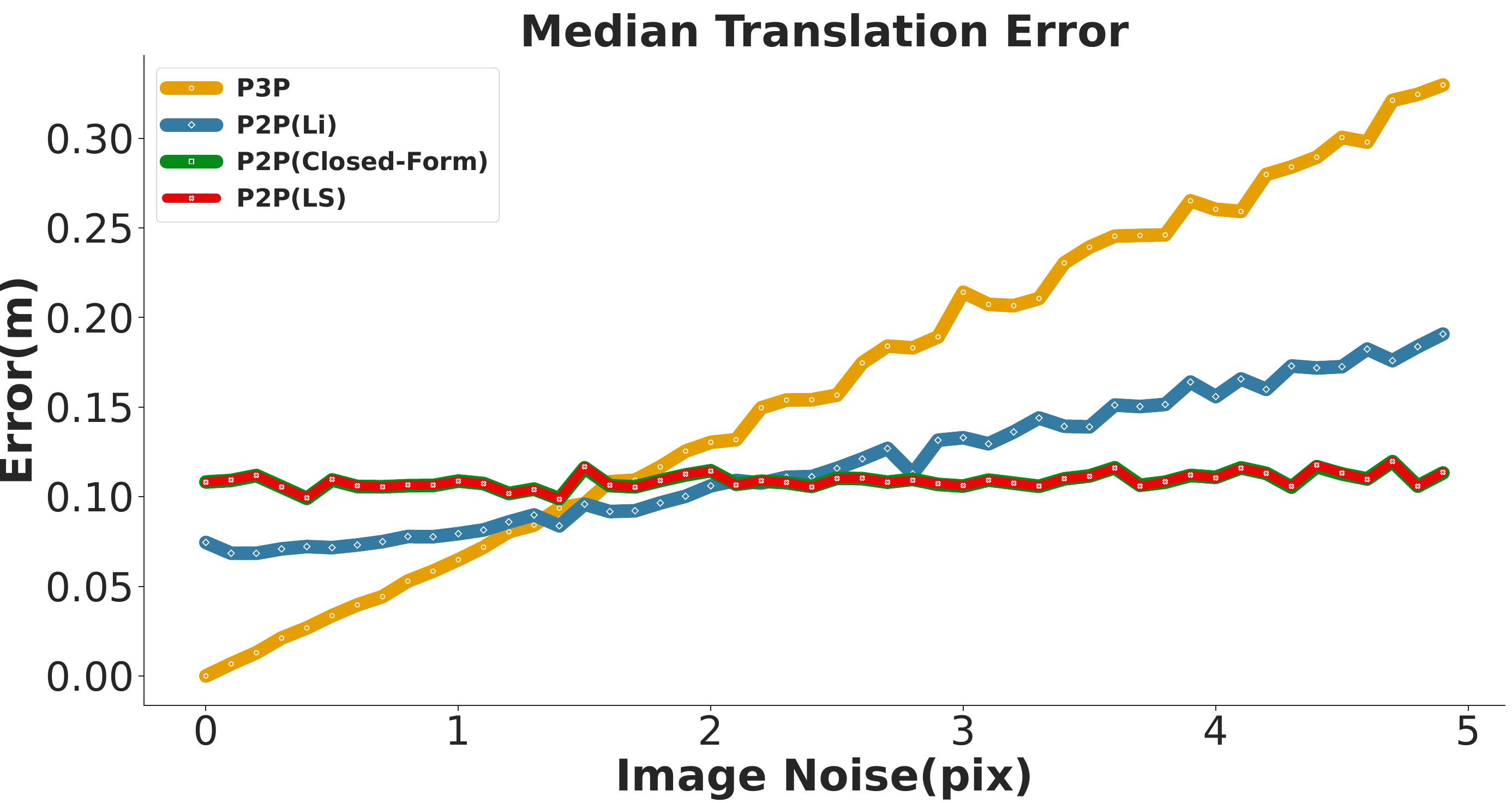}
            \includegraphics[width=0.49\linewidth]{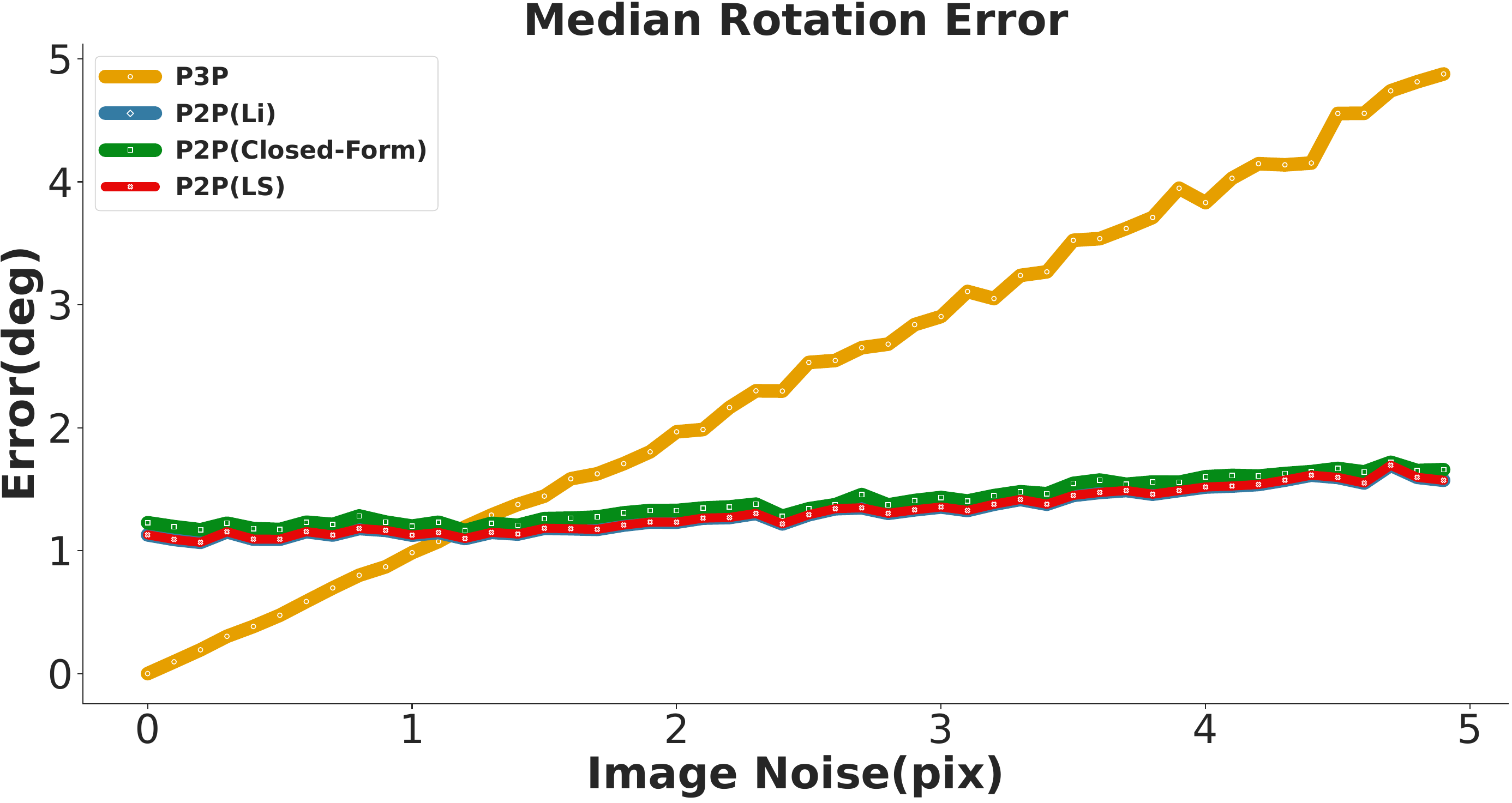}
        \label{fig:synthetic_px}
    }
    
    \hrulefill
    
    \subfigure[Median translation (m) and rotation (deg) errors under different target distances.]{ 
            \includegraphics[width=0.49\linewidth]{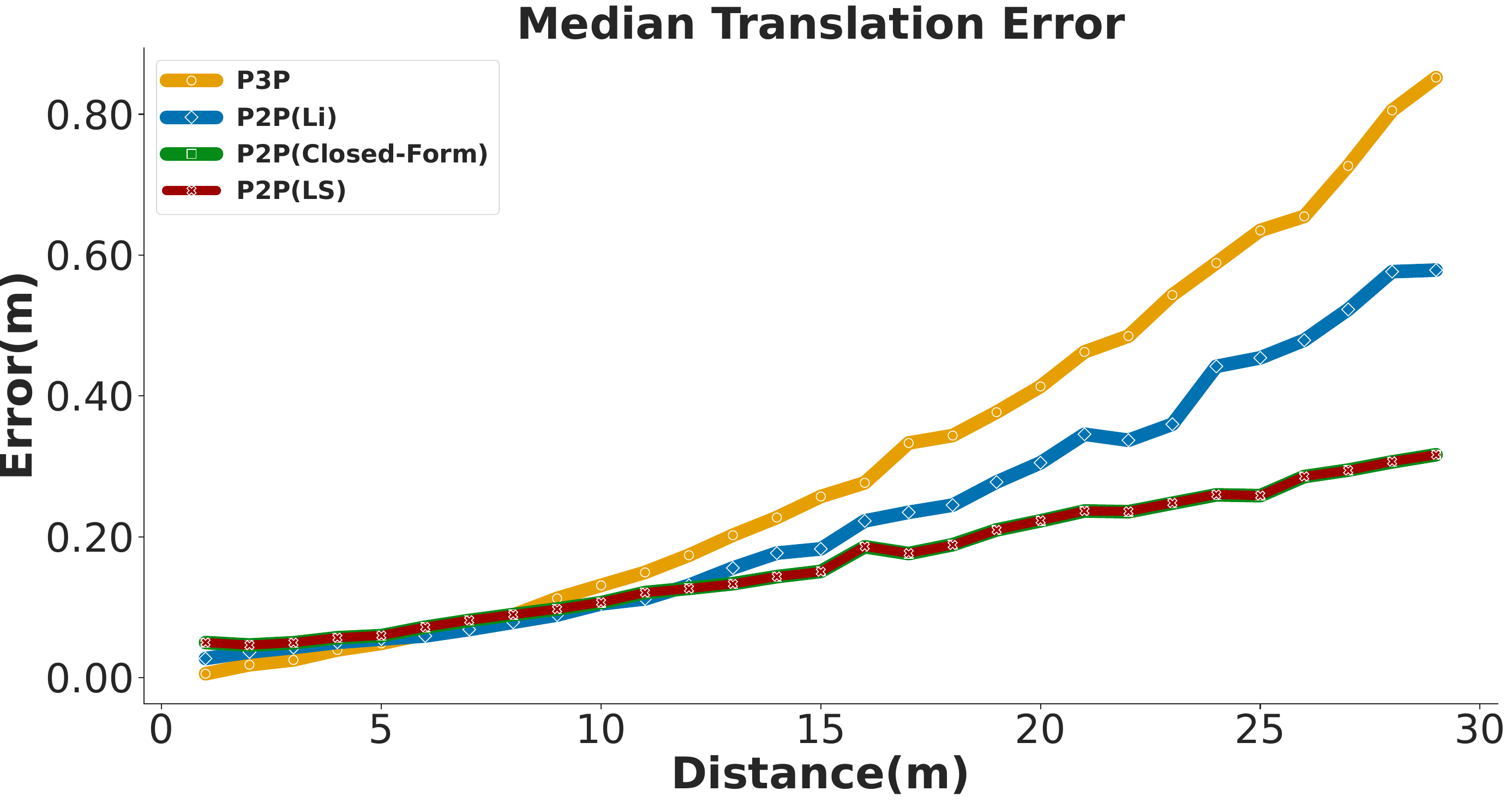}
            \includegraphics[width=0.49\linewidth]{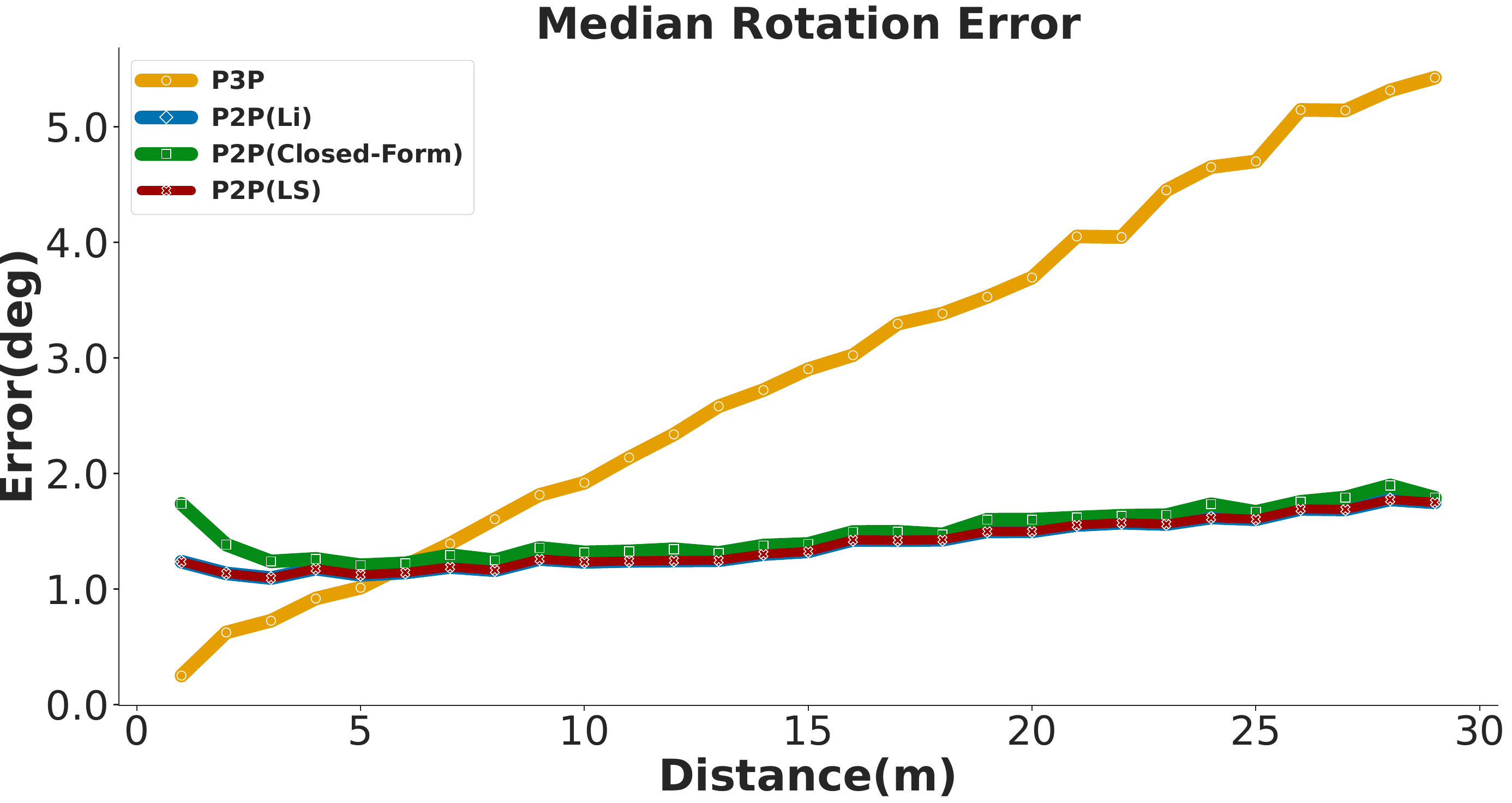}
        \label{fig:synthetic_distance}
    }%
    \caption{Translation and rotation errors with respect to the ground truth pose under different target distances and varying pixel noise levels. The proposed two P2P variants perform similarly in translation, and the LS variant shows comparable rotation accuracy to~\cite{li2023generalized}. The proposed P2P solvers show slightly higher errors under short-range or low-noise conditions, but demonstrate improved robustness and superior performance as the distance or noise level increases.}
    \label{fig:synthetic}
\end{figure*}
For the synthetic evaluation, we systematically sampled camera poses on spherical surfaces centered at the target's centroid with varying radii, to simulate different camera-to-target distances. Each camera is oriented toward the target's center to ensure all three points remain within the field of view. To emulate realistic operating conditions, independent zero-mean Gaussian perturbations are added to the sampled poses, with standard deviations of 1$^{\circ}$ for roll and pitch, and 0.03~m for height.

In event-based pose estimation with active markers, the observations correspond to spatially-distributed event clusters rather than idealized point features. This spatial dispersion inherently introduces localization uncertainty in the extracted keypoints. To quantitatively assess the sensitivity of the proposed solver to such measurement uncertainty, we evaluated its robustness under varying levels of pixel noise. Specifically, 1000 camera poses are sampled on a spherical surface with a radius of 10~m, and zero-mean Gaussian noise with standard deviations ranging from 0 to 5 pixels was added to the 2D point measurements.

Fig.~\ref{fig:synthetic_px} presents the median translation and rotation errors under growing levels of pixel noise. As the noise magnitude increases, all P2P-based methods demonstrate improved robustness in rotation estimation compared to P3P. However, incorporating the height prior yields a rotation performance nearly identical to that of~\cite{li2023generalized}, indicating that the two methods behave similarly for rotation estimation. As far as translation estimation is concerned, the closed-form and LS variants of the proposed method exhibit nearly identical performances, both consistently outperforming~\cite{kneip2011novel} and~\cite{li2023generalized} once the image noise becomes sufficiently large.

While the previous experiment focused on measurement noise, pose estimation accuracy is also affected by geometric factors, particularly the distance between the camera and the target. To analyze this effect, we fixed the pixel noise at $\sigma = 2$~px and sampled 1000 camera poses on spherical surfaces with radii ranging from~1 m to 30~m.

\begin{table}[b!]
\centering
\resizebox{\columnwidth}{!}{%
\begin{tabular}{|c|c|ccc|ccc|}
\hline
\multirow{2}{*}{Distance (m)} &
  \multirow{2}{*}{Method} &
  \multicolumn{3}{c|}{Position Error (m)} &
  \multicolumn{3}{c|}{Orientation Error (deg.)} \\ \cline{3-8} 
 &
   &
  \multicolumn{1}{c|}{Median} &
  \multicolumn{1}{c|}{Mean} &
  Std &
  \multicolumn{1}{c|}{Median} &
  \multicolumn{1}{c|}{Mean} &
  Std \\ \hline\hline
\multirow{4}{*}{5} &
  P3P~\cite{kneip2011novel} &
  \multicolumn{1}{c|}{\textbf{0.021}} &
  \multicolumn{1}{c|}{\textbf{0.025}} &
  \textbf{0.016} &
  \multicolumn{1}{c|}{\textbf{0.692}} &
  \multicolumn{1}{c|}{\textbf{0.791}} &
  \textbf{0.479} \\ \cline{2-8} 
 &
  P2P~\cite{li2023generalized} &
  \multicolumn{1}{c|}{0.037} &
  \multicolumn{1}{c|}{0.059} &
  0.069 &
  \multicolumn{1}{c|}{1.114} &
  \multicolumn{1}{c|}{1.264} &
  0.744 \\ \cline{2-8} 
 &
  P2P (Closed Form) &
  \multicolumn{1}{c|}{0.061} &
  \multicolumn{1}{c|}{0.083} &
  0.077 &
  \multicolumn{1}{c|}{1.191} &
  \multicolumn{1}{c|}{1.454} &
  1.086 \\ \cline{2-8} 
 &
  P2P (LS) &
  \multicolumn{1}{c|}{0.061} &
  \multicolumn{1}{c|}{0.083} &
  0.078 &
  \multicolumn{1}{c|}{1.116} &
  \multicolumn{1}{c|}{1.263} &
  0.755 \\ \hline\hline
\multirow{4}{*}{10} &
  P3P~\cite{kneip2011novel} &
  \multicolumn{1}{c|}{\textbf{0.078}} &
  \multicolumn{1}{c|}{\textbf{0.096}} &
  \textbf{0.072} &
  \multicolumn{1}{c|}{\textbf{1.430}} &
  \multicolumn{1}{c|}{\textbf{1.615}} &
  \textbf{0.976} \\ \cline{2-8} 
 &
  P2P~\cite{li2023generalized} &
  \multicolumn{1}{c|}{0.096} &
  \multicolumn{1}{c|}{0.143} &
  0.160 &
  \multicolumn{1}{c|}{1.216} &
  \multicolumn{1}{c|}{1.377} &
  0.815 \\ \cline{2-8} 
 &
  P2P (Closed Form) &
  \multicolumn{1}{c|}{0.109} &
  \multicolumn{1}{c|}{0.158} &
  0.155 &
  \multicolumn{1}{c|}{1.284} &
  \multicolumn{1}{c|}{1.565} &
  1.143 \\ \cline{2-8} 
 &
  P2P (LS) &
  \multicolumn{1}{c|}{0.109} &
  \multicolumn{1}{c|}{0.158} &
  0.155 &
  \multicolumn{1}{c|}{1.228} &
  \multicolumn{1}{c|}{1.370} &
  0.791 \\ \hline\hline
\multirow{4}{*}{20} &
  P3P~\cite{kneip2011novel} &
  \multicolumn{1}{c|}{0.302} &
  \multicolumn{1}{c|}{0.376} &
  0.296 &
  \multicolumn{1}{c|}{2.942} &
  \multicolumn{1}{c|}{3.267} &
  1.835 \\ \cline{2-8} 
 &
  P2P~\cite{li2023generalized} &
  \multicolumn{1}{c|}{0.313} &
  \multicolumn{1}{c|}{0.426} &
  0.416 &
  \multicolumn{1}{c|}{1.509} &
  \multicolumn{1}{c|}{1.691} &
  0.963 \\ \cline{2-8} 
 &
  P2P (Closed Form) &
  \multicolumn{1}{c|}{\textbf{0.217}} &
  \multicolumn{1}{c|}{\textbf{0.325}} &
  \textbf{0.329} &
  \multicolumn{1}{c|}{1.601} &
  \multicolumn{1}{c|}{1.901} &
  1.324 \\ \cline{2-8} 
 &
  P2P (LS) &
  \multicolumn{1}{c|}{\textbf{0.217}} &
  \multicolumn{1}{c|}{\textbf{0.325}} &
  \textbf{0.329} &
  \multicolumn{1}{c|}{\textbf{1.505}} &
  \multicolumn{1}{c|}{\textbf{1.678}} &
  \textbf{0.961} \\ \hline\hline
\multirow{4}{*}{30} &
  P3P~\cite{kneip2011novel} &
  \multicolumn{1}{c|}{0.684} &
  \multicolumn{1}{c|}{0.862} &
  0.723 &
  \multicolumn{1}{c|}{4.428} &
  \multicolumn{1}{c|}{4.976} &
  2.912 \\ \cline{2-8} 
 &
  P2P~\cite{li2023generalized} &
  \multicolumn{1}{c|}{0.658} &
  \multicolumn{1}{c|}{0.868} &
  0.828 &
  \multicolumn{1}{c|}{1.797} &
  \multicolumn{1}{c|}{2.041} &
  1.239 \\ \cline{2-8} 
 &
  P2P (Closed Form) &
  \multicolumn{1}{c|}{\textbf{0.327}} &
  \multicolumn{1}{c|}{\textbf{0.476}} &
  \textbf{0.475} &
  \multicolumn{1}{c|}{1.878} &
  \multicolumn{1}{c|}{2.254} &
  1.570 \\ \cline{2-8} 
 &
  P2P (LS) &
  \multicolumn{1}{c|}{\textbf{0.327}} &
  \multicolumn{1}{c|}{\textbf{0.476}} &
  \textbf{0.475} &
  \multicolumn{1}{c|}{\textbf{1.794}} &
  \multicolumn{1}{c|}{\textbf{2.027}} &
  \textbf{1.224} \\ \hline
\end{tabular}%
}
\caption{Quantitative results under different target distances. Best~values are bold.}
\label{tab:synthetic_distance}
\vspace{-8pt}
\end{table}

\begin{figure*}[t!]
    \centering
    \vspace{5pt}
    \subfigure[Pose estimation results under the degenerate configuration where two marker points have identical height (3 m).]{ 
            \includegraphics[width=0.49\textwidth]{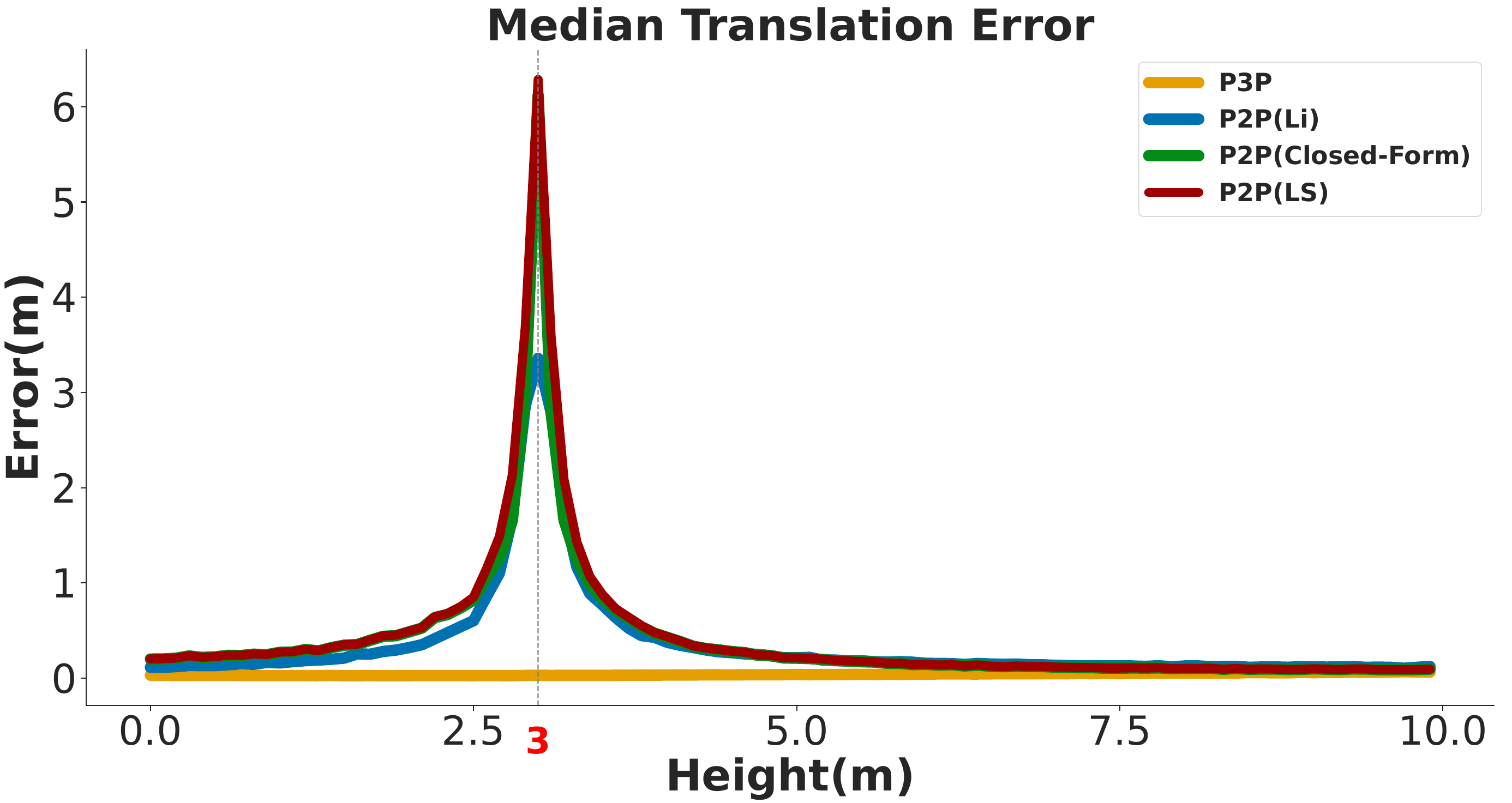}
            \includegraphics[width=0.49\textwidth]{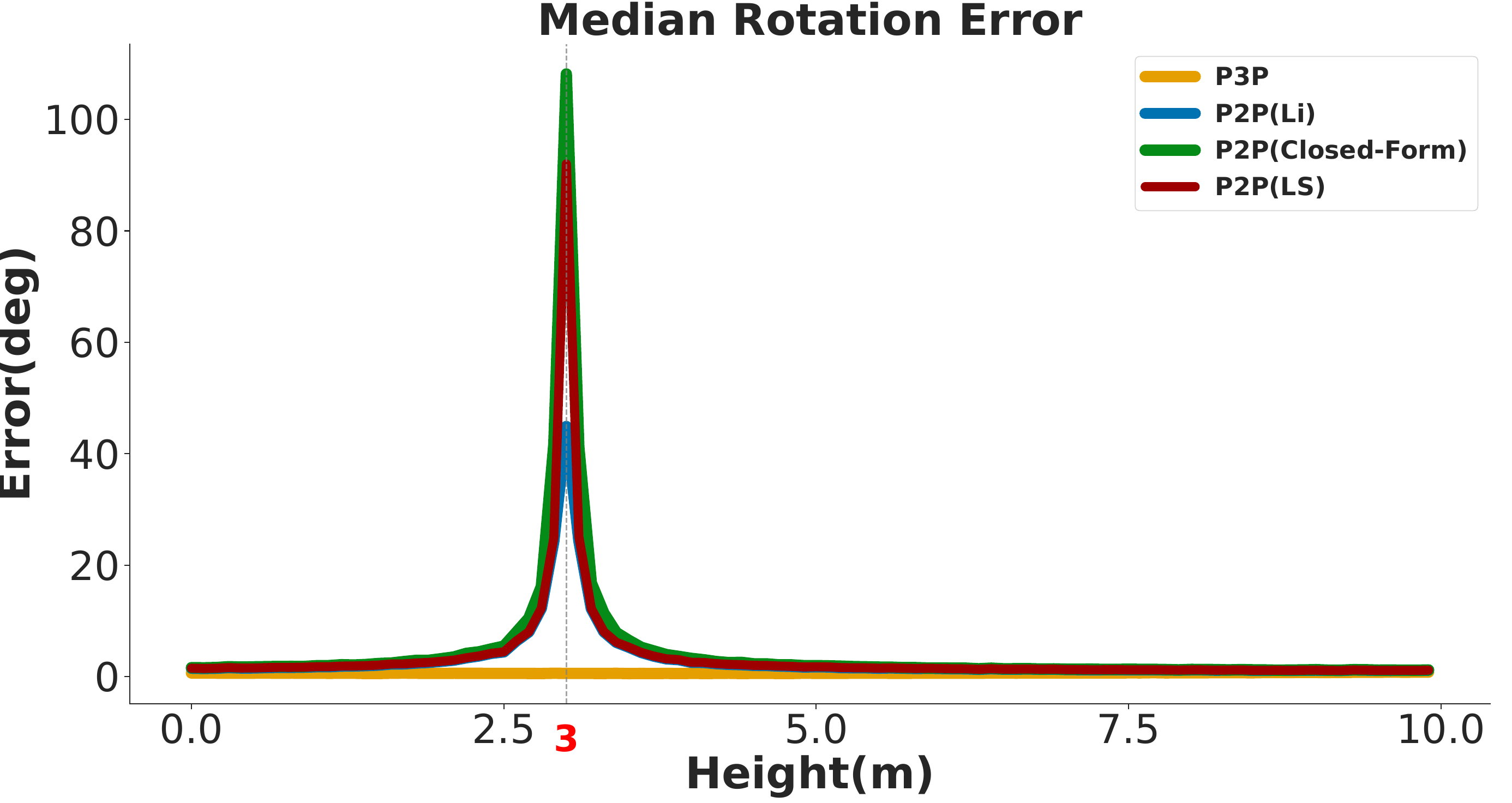}
        \label{fig:degenerate_same_height}
    }
    
    \hrulefill
    
    \subfigure[Pose estimation results under the degenerate configuration where two marker points have different height (3 m and 4 m).]{ 
            \includegraphics[width=0.49\textwidth]{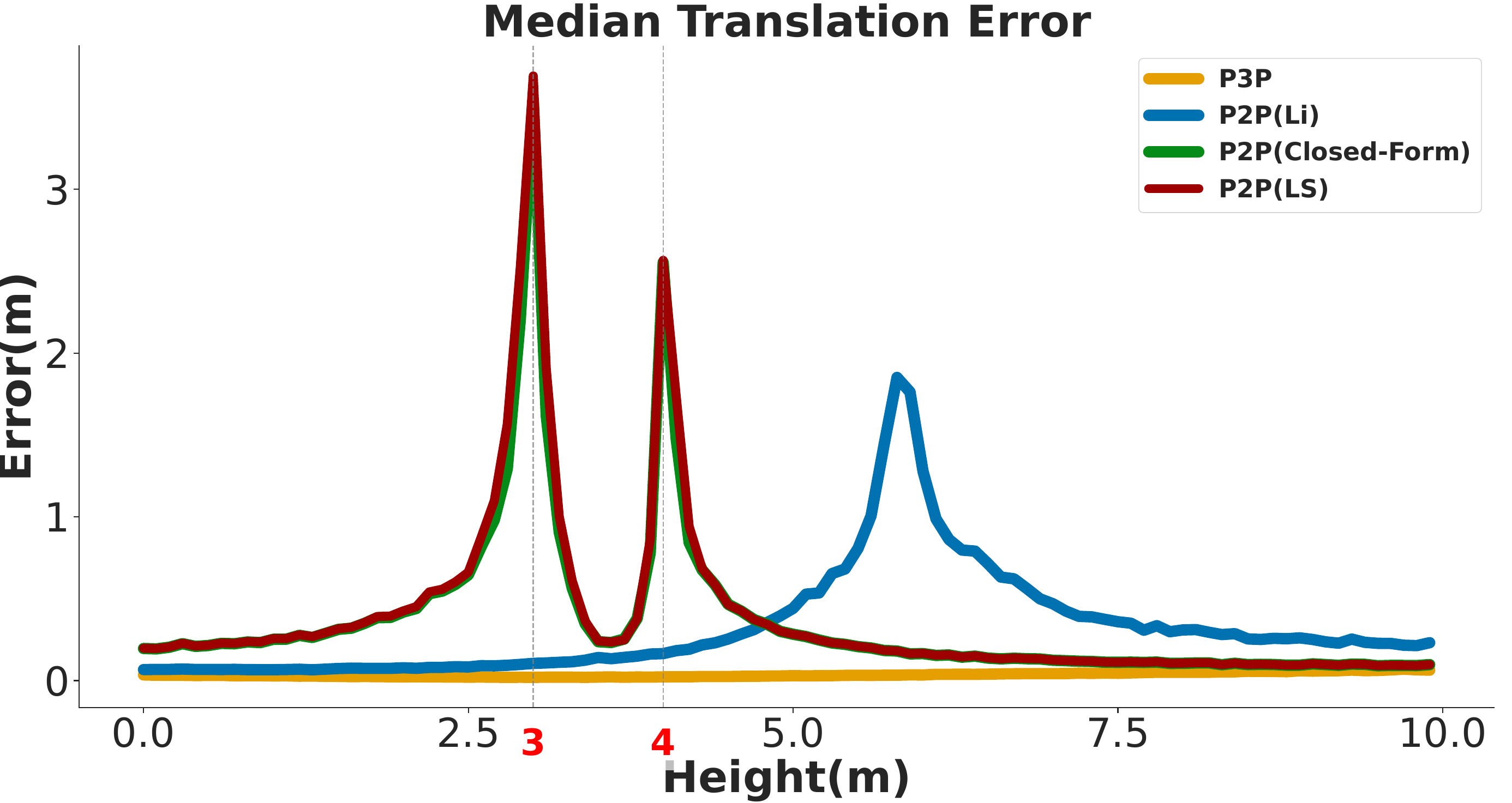}
            \includegraphics[width=0.49\textwidth]{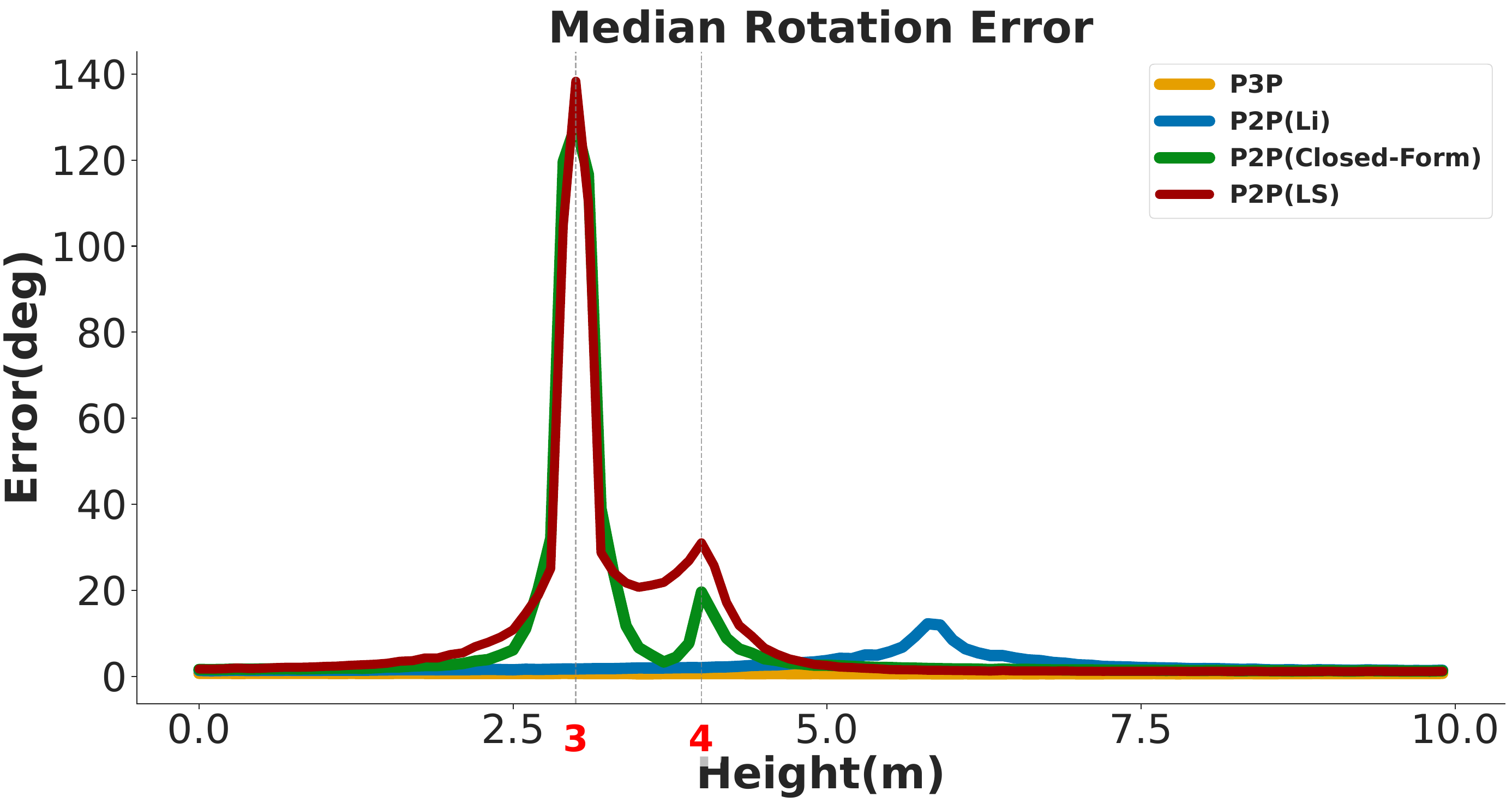}
        \label{fig:degenerate_diff_height}
    }%
    \caption{Pose estimation performance as a function of camera height. The top row corresponds to the configuration where the two marker points share identical height (3 m). In this case, all P2P-based methods become degenerate at a camera height of 3 m, exhibiting severe numerical instability.  The bottom row represents the configuration where the two marker points lie at different heights (3 m and 4 m).  The proposed P2P methods exhibit degeneracy only when the camera height approaches either point height, consistent with the theoretical analysis, whereas the method of~\cite{li2023generalized} shows degenerate behavior in configurations not explicitly discussed in the original paper. In contrast, P3P remains well-conditioned under all tested configurations. }
    \label{fig:degenerate}
\end{figure*}

Fig.~\ref{fig:synthetic_distance} presents the median translation and rotation errors across varying target distances, while Table~\ref{tab:synthetic_distance} summarizes the quantitative results. At short distances, the proposed P2P method achieves comparable translation and rotation accuracy to the approach of \cite{li2023generalized}, while all P2P-based methods underperform compared to P3P.
As the target distance increases, however, the P2P-based methods exhibit superior robustness in rotation estimation relative to P3P. For translation estimation, both variants of the proposed solver consistently yield lower errors than \cite{kneip2011novel} and \cite{li2023generalized} at larger distances.

In summary, in the synthetic evaluations, P2P-based methods demonstrated superior rotation robustness compared to P3P under image measurement uncertainty, whether induced by detection noise or increased viewing distance. Although the height prior has limited impact on rotation accuracy, it significantly improves translation robustness. The closed-form variant, however, is more sensitive to noise in rotation estimation than the LS formulation.

\subsection{Degeneracy Analysis}

We also conducted synthetic experiments to validate the degenerate configurations identified in the theoretical analysis. The trivial degenerate configuration in which the two 3D points differ only in height leads to a rank-deficient system with no solution, and it is therefore excluded from quantitative evaluation. We first considered the configuration where the two points lie at the same height (3~m) but differ in their horizontal coordinates. To facilitate the comparison with the P3P baseline, an additional non-collinear 3D point was introduced. Camera poses were sampled on horizontal planes at varying heights. For each height, 1000 poses were generated by applying random horizontal translations to ensure statistical generality. The same noise configuration as described in the previous subsection was applied.

Fig.~\ref{fig:degenerate_same_height} presents the median translation and rotation errors under this configuration. As expected, the solutions of the P3P solver are not degenerate in this scenario. 
In contrast, both the closed-form and LS variants of the proposed P2P method become numerically unstable as the camera height approaches that of the 3D points (3~m), which is consistent with our observability analysis. 
The method in~\cite{li2023generalized} exhibited a similar degenerate behavior. This aligns with their theoretical finding that the solver fails when both image rays become orthogonal to the rotation axis. Furthermore, outside the degenerate region, Li \emph{et al.}’s method achieves rotation accuracy identical to that of the LS variant of the proposed formulation, confirming the equivalence discussed in Sect.~\ref{Sect:TheorAnal}.

We further evaluated the case where the two 3D points lie at different heights (3~m and 4~m), while keeping the experimental setup unchanged. As shown in Fig.~\ref{fig:degenerate_diff_height}, both variants of the proposed method exhibit numerical instability when the camera's height approaches that of either 3D point, indicating that degeneracy arises whenever the vertical separation between the camera and one of the observed points vanishes.

Interestingly, \cite{li2023generalized} also exhibits degeneracy within a certain camera height range under this configuration, even though the corresponding image rays are not orthogonal to the rotation axis \cite{li2023generalized}. This phenomenon is not explicitly discussed in the original paper and suggests that additional geometric degeneracies might exist beyond those already identified by the authors. 

Overall, the degenerate cases are fully consistent with our theoretical analysis. 
With the height prior incorporated, degeneracy arises when the vertical configuration fails to provide sufficient geometric constraints, resulting in ill-conditioning and amplified noise sensitivity. These findings emphasize the importance of adequate height diversity between the camera and observed points to ensure stable pose estimation using the proposed solver.

\subsection{Validation on Real Data}
\subsubsection{Experiment Setup}
\begin{figure}[h!]
    \centering
    \subfigure{ 
            \includegraphics[width=0.49\linewidth]{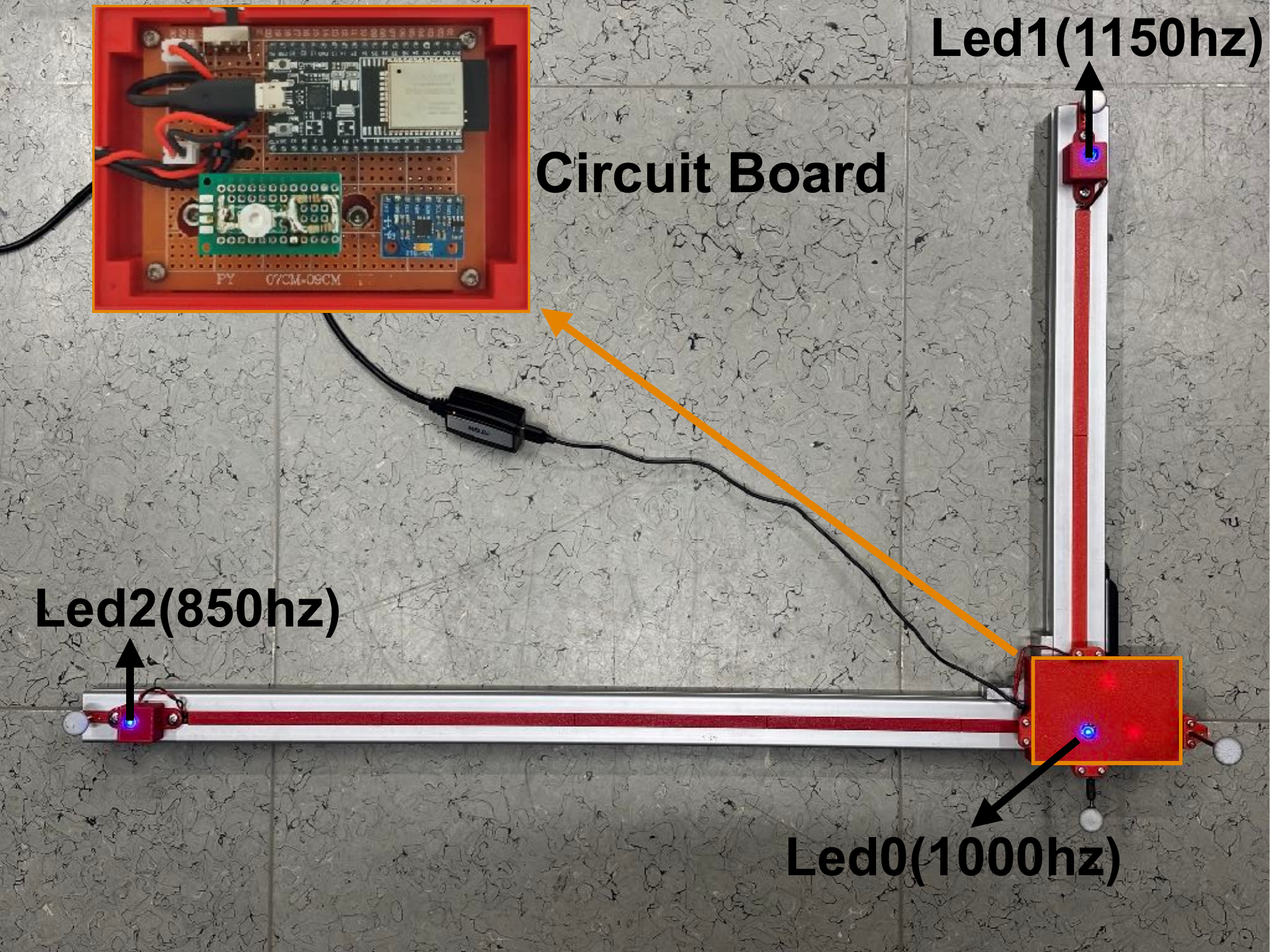}
            \includegraphics[width=0.49\linewidth]{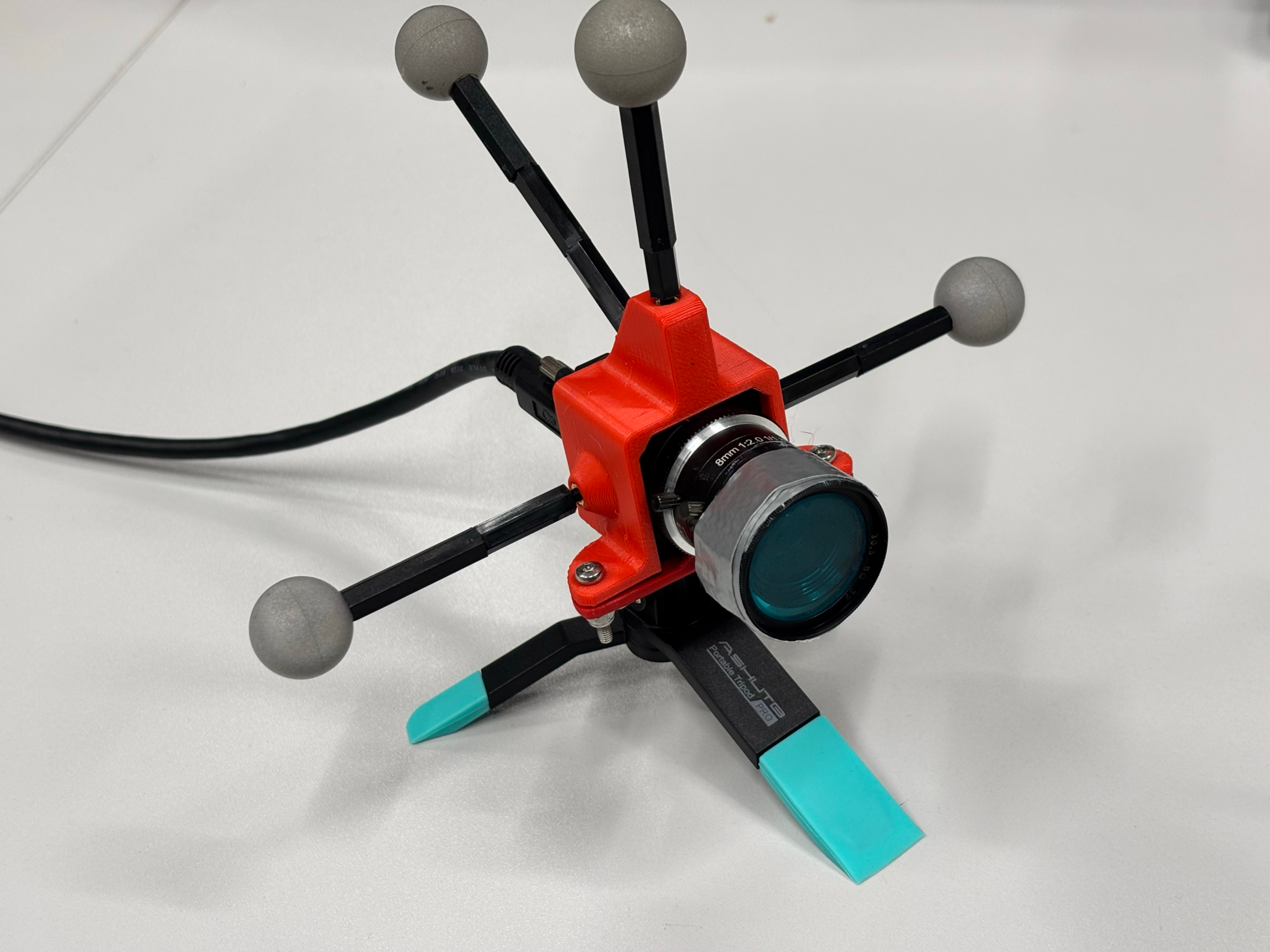}
    }%
    \caption{\emph{Real-world experimental setup}. (Left) LED-based target with active markers and (Right) EVK4 HD event camera.}
    \label{fig:led_evk}
    \vspace{-8pt}
\end{figure}

To demonstrate the practical applicability and robustness of the proposed method under real-world conditions, we carried out real-world experiments using a dedicated hardware platform (see Fig.~\ref{fig:led_evk}). 
The system is built upon an EVK4 HD camera from Prophesee, which incorporates an IMX636ES event-based vision sensor with a resolution of 1280 $\times$ 720 pixels, together with a Soyo SFA0820-5M lens. As the active target, three high-power LEDs with broad emission angles are rigidly mounted on a custom-designed structure to ensure fixed spatial geometry. The LEDs are driven by an ESP32 microcontroller using PWM signals to generate stable and frequency-controlled blinking patterns. The operating frequencies are set to 1000~Hz, 1150~Hz, and 850~Hz, respectively, to ensure sufficient frequency spacing for reliable separation in the event stream.

The ground-truth camera poses and the 3D positions of the LED markers were obtained with an OptiTrack motion capture system. The event camera was intrinsically calibrated using the Prophesee SDK with a blinking chessboard pattern. In addition, hand–eye calibration~\cite{daniilidis1999hand} between the event camera and the OptiTrack coordinate frame was performed using the OpenCV library. 
The positions of 3D LED markers, together with their associated blinking frequencies, were provided to the localization system as prior information. The ground-truth tilt angle was given to \cite{li2023generalized}, while the proposed P2P solvers were provided with the ground-truth height.

\subsubsection{LED Marker Detection}
In order to detect frequency-modulated LED markers in the event stream, we designed a simple two-stage clustering-based pipeline, operating in the spatial and temporal domains. Specifically, events are accumulated within a 20~ms time window to ensure sufficient signal density. Within this window, spatial clustering is performed in the image plane using DBSCAN~\cite{SchubertSaEsKrXu_TDS17} to identify candidate LED regions. This step forms spatial clusters for each LED and removes scattered noise events. For each spatial cluster, a further temporal clustering step is performed to estimate the blinking frequency. By~examining the temporal distribution of events, the dominant modulation frequency associated with each cluster is determined. Finally, for each detected frequency, the pixel coordinates of the associated events are averaged to produce a single observation point. The resulting frequency-labeled pixel locations are then used as the input measurements for pose estimation.

\subsubsection{Real-World Experimental Results}
\begin{figure}[t!]
    \centering
    \vspace{5pt}
    \subfigure{ 
        \begin{minipage}[b]{\linewidth}
            \centering
            \includegraphics[width=0.49\linewidth]{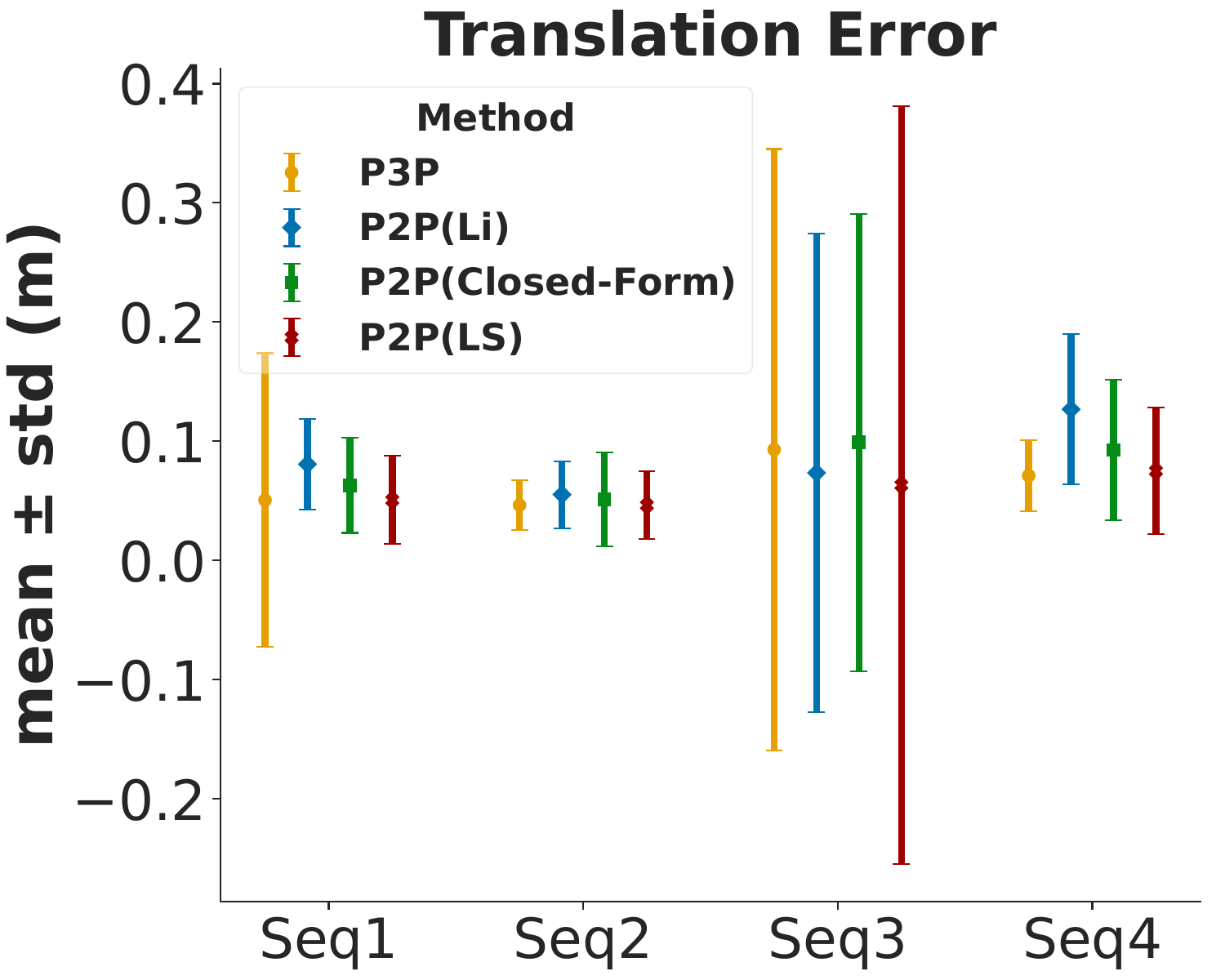}
            \includegraphics[width=0.49\linewidth]{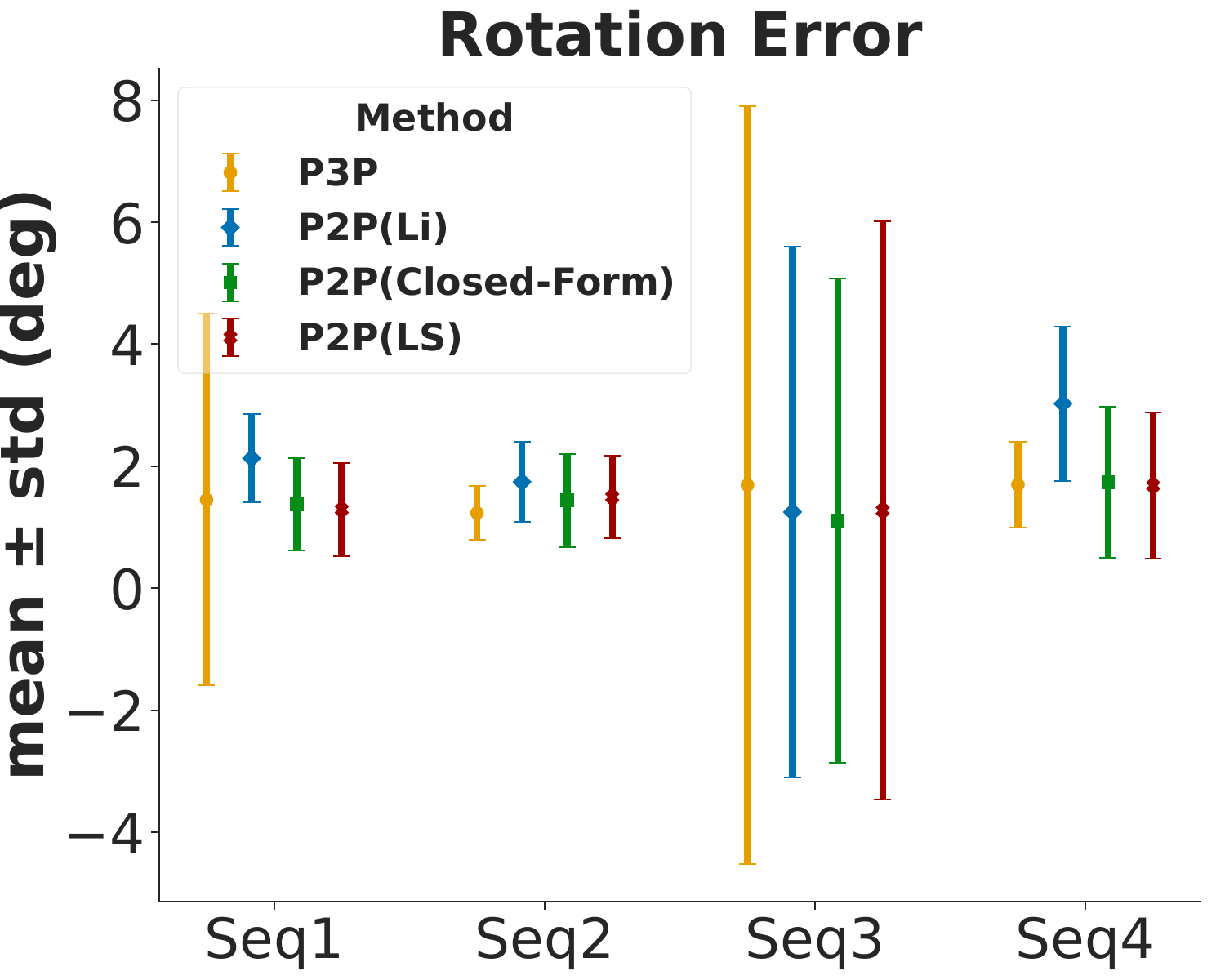}
        \end{minipage}}%
    \caption{Translation and rotation error distribution in the real-world experiments. The mean and standard deviation of the error are presented for all methods over the four sequences.}
    \label{fig:real_bar}
\end{figure}


\newcommand{\graycell}[1]{\cellcolor{gray!10}#1}
\begin{table}[t!]
\resizebox{\columnwidth}{!}{%
\begin{tabular}{|c|c|ccc|ccc|}
\hline
\multirow{2}{*}{Sequence} &
  \multirow{2}{*}{Method} &
  \multicolumn{3}{c|}{Position Error (m)} &
  \multicolumn{3}{c|}{Orientation Error (deg.)} \\ \cline{3-8} 
 &
   &
  \multicolumn{1}{c|}{Median} &
  \multicolumn{1}{c|}{Mean} &
  Std &
  \multicolumn{1}{c|}{Median} &
  \multicolumn{1}{c|}{Mean} &
  Std \\ \hline\hline
\multirow{4}{*}{Seq. 1} &
  \graycell{P3P~\cite{kneip2011novel}} &
  \multicolumn{1}{c|}{\graycell{0.035}} &
  \multicolumn{1}{c|}{\graycell{0.051}} &
  \graycell{0.123} &
  \multicolumn{1}{c|}{\graycell{1.106}} &
  \multicolumn{1}{c|}{\graycell{1.450}} &
  \graycell{3.046} \\ \cline{2-8} 
 &
  P2P~\cite{li2023generalized} &
  \multicolumn{1}{c|}{0.075} &
  \multicolumn{1}{c|}{0.081} &
  0.038 &
  \multicolumn{1}{c|}{2.046} &
  \multicolumn{1}{c|}{2.130} &
  \textbf{0.727} \\ \cline{2-8} 
 &
  P2P (Closed Form) &
  \multicolumn{1}{c|}{0.055} &
  \multicolumn{1}{c|}{0.063} &
  0.040 &
  \multicolumn{1}{c|}{1.335} &
  \multicolumn{1}{c|}{1.377} &
  0.753 \\ \cline{2-8} 
 &
  P2P (LS) &
  \multicolumn{1}{c|}{\textbf{0.043}} &
  \multicolumn{1}{c|}{\textbf{0.051}} &
  \textbf{0.037} &
  \multicolumn{1}{c|}{\textbf{1.220}} &
  \multicolumn{1}{c|}{\textbf{1.289}} &
  0.765 \\ \hline\hline
\multirow{4}{*}{Seq. 2} &
  \graycell{P3P~\cite{kneip2011novel}} &
  \multicolumn{1}{c|}{\graycell{0.041}} &
  \multicolumn{1}{c|}{\graycell{0.046}} &
  \graycell{0.021} &
  \multicolumn{1}{c|}{\graycell{1.184}} &
  \multicolumn{1}{c|}{\graycell{1.234}} &
  \graycell{0.441} \\ \cline{2-8} 
 &
  P2P~\cite{li2023generalized} &
  \multicolumn{1}{c|}{0.050} &
  \multicolumn{1}{c|}{0.055} &
  \textbf{0.028} &
  \multicolumn{1}{c|}{1.710} &
  \multicolumn{1}{c|}{1.742} &
  \textbf{0.652} \\ \cline{2-8} 
 &
  P2P (Closed Form) &
  \multicolumn{1}{c|}{0.042} &
  \multicolumn{1}{c|}{0.051} &
  0.039 &
  \multicolumn{1}{c|}{\textbf{1.412}} &
  \multicolumn{1}{c|}{\textbf{1.442}} &
  0.762 \\ \cline{2-8} 
 &
  P2P (LS) &
  \multicolumn{1}{c|}{\textbf{0.041}} &
  \multicolumn{1}{c|}{\textbf{0.046}} &
  \textbf{0.028} &
  \multicolumn{1}{c|}{1.438} &
  \multicolumn{1}{c|}{1.494} &
  0.678 \\ \hline\hline
\multirow{4}{*}{\begin{tabular}[c]{@{}c@{}}Seq. 3\\ (Target distance:\\1.25~-~2.75~m)\end{tabular}} &
  \graycell{P3P~\cite{kneip2011novel}} &
  \multicolumn{1}{c|}{\graycell{0.053}} &
  \multicolumn{1}{c|}{\graycell{0.083}} &
  \graycell{0.211} &
  \multicolumn{1}{c|}{\graycell{1.209}} &
  \multicolumn{1}{c|}{\graycell{1.932}} &
  \graycell{5.049} \\ \cline{2-8} 
 &
  P2P~\cite{li2023generalized} &
  \multicolumn{1}{c|}{0.062} &
  \multicolumn{1}{c|}{0.076} &
  0.102 &
  \multicolumn{1}{c|}{1.637} &
  \multicolumn{1}{c|}{1.919} &
  2.088 \\ \cline{2-8} 
 &
  P2P (Closed Form) &
  \multicolumn{1}{c|}{\textbf{0.050}} &
  \multicolumn{1}{c|}{\textbf{0.065}} &
  \textbf{0.098} &
  \multicolumn{1}{c|}{1.511} &
  \multicolumn{1}{c|}{1.746} &
  1.923 \\ \cline{2-8} 
 &
  P2P (LS) &
  \multicolumn{1}{c|}{0.051} &
  \multicolumn{1}{c|}{0.066} &
  0.107 &
  \multicolumn{1}{c|}{\textbf{1.477}} &
  \multicolumn{1}{c|}{\textbf{1.686}} &
  \textbf{2.021} \\ \hline\hline
\multirow{4}{*}{\begin{tabular}[c]{@{}c@{}}Seq. 4\\ (Target distance:\\3.4~-~4.6~m)\end{tabular}} &
  \graycell{P3P~\cite{kneip2011novel}} &
  \multicolumn{1}{c|}{\graycell{0.066}} &
  \multicolumn{1}{c|}{\graycell{0.071}} &
  \graycell{0.030} &
  \multicolumn{1}{c|}{\graycell{1.563}} &
  \multicolumn{1}{c|}{\graycell{1.698}} &
  \graycell{0.700} \\ \cline{2-8} 
 &
  P2P~\cite{li2023generalized} &
  \multicolumn{1}{c|}{0.113} &
  \multicolumn{1}{c|}{0.127} &
  0.063 &
  \multicolumn{1}{c|}{2.876} &
  \multicolumn{1}{c|}{3.026} &
  1.266 \\ \cline{2-8} 
 &
  P2P (Closed Form) &
  \multicolumn{1}{c|}{0.078} &
  \multicolumn{1}{c|}{0.093} &
  0.059 &
  \multicolumn{1}{c|}{1.529} &
  \multicolumn{1}{c|}{1.739} &
  1.235 \\ \cline{2-8} 
 &
  P2P (LS) &
  \multicolumn{1}{c|}{\textbf{0.062}} &
  \multicolumn{1}{c|}{\textbf{0.075}} &
  \textbf{0.053} &
  \multicolumn{1}{c|}{\textbf{1.469}} &
  \multicolumn{1}{c|}{\textbf{1.681}} &
  \textbf{1.196} \\ \hline
\end{tabular}%
}
\caption{Quantitative results in the real-world experiments.}
\vspace{-8pt}
\label{tab:real}
\end{table}

In the real-world experiments, only $\mathrm{LED1}$ and $\mathrm{LED2}$ (see~Fig.~\ref{fig:led_evk}) are used in the P2P-based solver to avoid the same-height special configuration discussed in Sect~\ref{Sect:TheorAnal}, under which the two solvers become equivalent. In contrast to the proposed solver, the formulations in~\cite{kneip2011novel} and~\cite{li2023generalized} both admit multiple valid solutions. For a fair comparison, the solution yielding the lowest pose error with respect to the ground truth was selected for evaluation. Four trajectory sequences were collected for real-world evaluation. Seq.~1 and Seq.~2 involve arbitrary motions within the workspace to validate the feasibility of the proposed method in practical scenarios. Seq.~3 and Seq.~4 correspond to motions performed in relatively near and far regions, respectively, to assess the performance of the proposed approach at different target distances.

Fig.~\ref{fig:real_bar} shows the translation and rotation error distribution across all sequences,
while Table~\ref{tab:real} reports the corresponding quantitative statistics in detail. As P3P operates under a stronger three-point assumption, it is shaded in gray as a reference baseline, while the primary evaluation is performed within the two-point setting. Boldface indicates the best performance among the P2P methods. As reported in Table~\ref{tab:real}, P3P achieves the lowest mean rotation error in most sequences, while its translation performance is competitive, but not consistently superior to that of the proposed P2P solver. However, a relatively larger standard deviation is observed in certain cases, suggesting increased sensitivity to measurement noise and occasional false detections. In~contrast, the proposed method exhibits consistently lower variance, indicating improved numerical stability and robustness. 

Within the P2P framework, the proposed solver consistently outperforms~\cite{li2023generalized} in both translation and rotation accuracy. The improvement is particularly evident in the far-range sequence (Seq. 4), where the estimation is more sensitive to measurement noise. Both the LS and closed-form variants achieve comparable performance, with the LS variant attaining slightly lower errors and exhibiting improved robustness.

Overall, the incorporation of height information enhances the robustness of the proposed P2P formulation, particularly in translation estimation.
Moreover, unlike multi-solution minimal solvers, such as~\cite{kneip2011novel,li2023generalized}, the proposed approach delivers a unique solution, which makes it highly attractive in real-world applications. 


\section{CONCLUSION}\label{Sect:Concl}

In this paper, we presented a robust and accurate 2-point solver for event-based pose estimation from active LED markers. We formulated the P2P problem with an extra prior on the camera height, derived a closed-form and a least-squares solution, and conducted a rigorous theoretical analysis, which shed light on the properties of the proposed solver. We performed extensive evaluations on both synthetic and real data, including a detailed degeneracy study. The results indicate that our solver, with height prior, provides a unique solution with better estimates of the camera's translation, 
and that it is robust against large viewing distances from the markers. Overall, it offers improved accuracy over the state-of-the-art P2P solvers~\cite{li2023generalized}, and it remains competitive with the P3P baseline~\cite{kneip2011novel}.


\section{ACKNOWLEDGEMENT}

This work was supported by the French and Austrian National Research Agencies (FWF, ANR) through the EVELOC project (ANR-23-CE33-0011, I 6747-N), 2024-2028.

\balance
\bibliographystyle{IEEEtran}
\bibliography{IEEEexample}
\end{document}